# Extracting Man-Made Objects From Remote Sensing Images via Fast Level Set Evolutions

Zhongbin Li, Wenzhong Shi, Qunming Wang, and Zelang Miao

***Abstract***—Object extraction from remote sensing images has long been an intensive research topic in the field of surveying and mapping. Most existing methods are devoted to handling just one type of object and little attention has been paid to improving the computational efficiency. In recent years, level set evolution (LSE) has been shown to be very promising for object extraction in the community of image processing and computer vision because it can handle topological changes automatically while achieving high accuracy. However, the application of state-of-the-art LSEs is compromised by laborious parameter tuning and expensive computation. In this paper, we proposed two fast LSEs for man-made object extraction from high spatial resolution remote sensing images. The traditional mean curvature-based regularization term is replaced by a Gaussian kernel and it is mathematically sound to do that. Thus a larger time step can be used in the numerical scheme to expedite the proposed LSEs. In contrast to existing methods, the proposed LSEs are significantly faster. Most importantly, they involve much fewer parameters while achieving better performance. The advantages of the proposed LSEs over other state-of-the-art approaches have been verified by a range of experiments.

## ACRONYMS

| | |
|---|---|
| B | blue; |
| BF | binary function; |
| Caltech | California Institute of Technology; |
| CV | Chan and Vese; |
| DRLSE | distance regularized level set evolution; |
| $E_{T_d}$ | edge-based data term; |
| G | green; |
| GAC | geodesic active contour; |
| GIS | geographic information system; |
| JPL | Jet Propulsion Laboratory; |
| LSE | level set evolution; |
| LSF | level set function; |
| NDVI | normalized difference vegetation index; |
| NIR | near-infrared; |
| R | red; |
| $R_{T_d}$ | region-based data term; |
| SAR | synthetic aperture radar; |
| SDF | signed distance function; |
| ZLC | zero level curve. |

This work was supported in part by Ministry of Science and Technology of China (Project No. 2012AA12A305).

Z.B. Li, W.Z. Shi, and Q.M. Wang are with the Department of Land Surveying and Geo-Informatics, The Hong Kong Polytechnic University, Kowloon 999077, Hong Kong.

Z.L. Miao is with the Department of Land Surveying and Geo-informatics, The Hong Kong Polytechnic University, Hong Kong, and also with the School of Environmental Science and Spatial Informatics, China University of Mining and Technology, Xuzhou, Jiangsu 221116, China.

(e-mail: lzbtongji@gmail.com; lswzshi@polyu.edu.hk; wqm11111@126.com; cumtzlmiao@gmail.com).

## I. INTRODUCTION

Extracting desired information from remote sensing data has been an intensive research topic in surveying and mapping for decades. In particular, extracting man-made objects (e.g., road networks, building roofs, and airport runways) from high spatial resolution optical images is beneficial for a diverse range of applications such as quantification of impervious surfaces [1-3], thematic cartography [4], timely update of urban geographic information system (GIS) [5, 6], disaster assessment [7, 8], and military reconnaissance [9]. Typically, man-made objects can be identified by using their intrinsic geometric properties or spectral signatures [10]. For instance, roads often appear as elongated features with homogeneous intensities [11-13], and thus, extracting road networks in some sense amounts to detecting line segments. Different from roads, building roofs generally consist of rectangles [14] or regular polygons with parallel lines [15] and right angles. As a consequence, the corner detector is often employed for building extraction in some studies [16, 17]. In addition, building shadows generated due to the slanted incident light in remote sensing images can also provide auxiliary information for building extraction [18].

In hyperspectral images, the abundant spectral information is useful for image classification [19, 20], subpixel target detection [21], and extraction of natural objects (e.g., vegetation, bare soil, natural water body, and shoreline) [10]. However, despite the rapid advancement of remote sensing technology, the available spectral information for the commonly used high spatial resolution optical images is very limited. For example, only three visible bands (i.e., blue (B), green (G), red (R)) and one near-infrared (NIR) band are currently available for images offered by satellite sensors such as Ikonos, Quickbird, Pleiades-1, and Geoeye-1. In addition, Worldview-1 only has a panchromatic band and no multispectral bands available. Although there are four more spectral bands available for Worldview-2, they are mainly used for the vegetation analysis. Using the limited spectral signature alone is not enough to extract man-made objects, especially in cases where the target objects (e.g., road networks, building roofs, and parking lots) are constructed with the same material (e.g., concrete or asphalt) and have similar spectral signatures.

Over the past few years, a series of approaches has been developed for object extraction from optical remote sensing images. Some comprehensive reviews can be found in [22, 23]. However, man-made object extraction from optical remote sensing images is still an open problem and the state-of-the-art methods face the following four major challenges.

1) **The complexity of optical images.** Today, the increasing spatial and spectral resolutions of optical remote sensing images lead to a rapid increase of complexity of image analysis correspondingly [24]. The complex scenes often pose great challenges for object extraction. That is because the geometric shapes, spectral signatures, and texture features of the backgrounds are highly similar to those of desired objects. In addition, despite the high performance computing, the challenges for fast processing of large volumes of remote sensing data still remain [25].

2) **Spectral limitation.** As mentioned before, there are only four spectral bands (i.e., B, G, R, and NIR) available for high spatial resolution images offered by sensors such as Ikonos, Quickbird, Pleiades-1, and Geoeye-1. Although Worldview-2 has four more spectral bands, they are primarily used for the vegetation analysis.

3) **Automation**. Currently, the degree of automation for object extraction from remote sensing images is still not very high. Actually, it is challenging to devise a fully automated system [26]. From a practical perspective, semiautomatic methods with appropriate human intervention are more appealing [23].

4) **Multi-object extraction.** Most state-of-the-art methods for object extraction from remote sensing images were developed to extract one specific type of object instead of multiple kinds of objects.

Based on the above analysis, we found that there is a need to develop more efficient techniques for object extraction, particularly for dealing with multiple objects. With this in mind, we proposed two fast level set evolutions (LSEs) for semiautomatic extraction of multiple kinds of man-made objects in this paper. Our essential contributions are as follows:

1) For applicability and efficiency, we make further development of traditional LSEs in three aspects. Specifically, we begin by eliminating the commonly used curvature-based regularization term from traditional LSEs. Instead, we employ a Gaussian kernel to keep the evolving LSEs smooth in a separate step. As pointed out in [27, 28], the convolution of a signal with a Gaussian kernel is equivalent to the solution of heat diffusion equation with the signal as the original value. Therefore, it is mathematically feasible to replace the curvature-based regularization term by a Gaussian kernel.

Then, we propose two fast LSEs: one is an edge-based LSE and the other is a region-based one. Compared with existing LSEs, they have much fewer parameters, thereby avoiding the labor intensive parameter tuning. Finally and most importantly, a larger time step can be used in the numerical scheme of the proposed LSEs, thus expediting the evolution substantially. All these improvements make the proposed LSEs more efficient.

2) The proposed LSEs are capable of extracting most man-made objects from optical remote sensing images by just using the gradient or region statistics (e.g., intensity mean or variance) of the objects instead of geometric shapes and spectral information. In this respect, they are more generic and operational than those methods that can only extract one specific kind of object.

3) We apply the proposed LSEs to extract multiple types of man-made objects, such as building roofs, road networks, and airport runways from a number of high spatial resolution optical images. To verify their advantages, we compare them with other state-of-the-art LSEs in terms of accuracy, parameter tuning, and computation time.

The study is organized as follows. In Section II, we begin with a brief review of the general approaches for man-made object extraction from remote sensing images. Then, we focus on the LSE due to its growing acceptance in the field of remote sensing. In Section III, we describe the proposed LSEs in detail. In Section IV, the proposed LSEs are employed to extract multiple types of man-made objects from different images and their characteristics are analyzed by comparison with other state-of-the-art LSEs. In Section V, the experimental results are further discussed and several perspectives for future research are offered. Finally, Section VI concludes the whole paper.

## II. Previous Work

In this section, we first give a brief review of the general approaches for man-made object extraction from remote sensing images. Next, we introduce the level set evolutions (LSEs) and their applications in remote sensing. Finally, we give a brief discussion about the disadvantages of the state-of-the-art methods. As the focus of this study is mainly on the extraction of man-made objects, the methods for natural object extraction are not included here.

### A. General Approaches for Man-Made Object Extraction

*1) Geometric Shapes (e.g., Line Segment, Rectangle, Right Angle, and Parallel Line):* In remote sensing images, roads and buildings are often composed of linear boundaries and homogenous regions with marked contrast to the nearby objects. Thus, line detection becomes one of the most commonly used methods for road [11, 29-31] and building detection [32]. A number of studies extracted building roofs by detecting parallel line segments such as [33] and [34]. In recent years, stochastic geometry models such as Gibbs and marked point process have also been utilized for man-made object extraction [35-38]. In a different context, local features such as L- and T-shape junctions [39], linear feature [40], and corners [16] are employed for road and building extraction.

*2) Spectral (Color) Information:* Spectral information is another main source for man-made object extraction. For instance, NDVI was used to remove the tree-generated lines for further refinement of the building extraction [41]. More recently, NDVI and brightness index were utilized to extract vegetation and shadows from Ikonos images, respectively [42]. In addition, the color space transformation was also used in some studies for object extraction from remote sensing images. For example, the original RGB images were converted into YIQ ones for building extraction in [41]. In [32], building shadows were extracted from HIS color space. Recently, the original RGB images were transformed into HSV and YIQ color spaces for building roof extraction [17].

*3) Shadow:* In some of the latest studies, the shadow is regarded as one of the important clues for building reconstruction. For instance, it was used to estimate the building height from satellite images [32], assess the damages caused by the earthquake by comparing the differences between the pre- and post-seismic building shadows [43], and produce labels for the semi-automated GrabCut approach for building extraction [18].

### B. Level Set Evolutions

The key idea of LSEs is to track the moving zero-level-set in a dynamic higher-dimensional level set function (LSF). The intersection between the zero-level-set and the LSF is called zero level curve (ZLC), which is strictly closed. Thus, tracking the movement of the zero-level-set is equivalent to describing the variation of ZLC. With the evolution of the LSF, ZLC changes automatically and it continues to move until some stopping criteria are met. We typically refer to the stopping criterion as the data term that pushes the moving ZLCs toward the desired object boundaries. During the evolution of ZLCs, some regularization terms are used to keep ZLCs smooth and regular. The original LSE proposed by Osher and Sethian [44] is given as follows:

$$\phi_t = F(\kappa)|\nabla\phi| \tag{1}$$

where $\phi$ is the LSF; $t$ is a temporal variable; $\kappa$ is the mean curvature of ZLC; $F(\kappa)$ is a function with respect to $\kappa$; and $\nabla$ is the gradient operator.

In order to obtain accurate and robust results, a variety of data terms and regularizing terms have been proposed within the past few years. According to the data terms, LSEs can be grouped into two classes: edge-based and region-based. In the following, we investigate the two types of LSEs in detail. Some review articles regarding LSEs can be found in [45, 46] for 2D object extraction and [47, 48] for 3D shape reconstruction.

### 1) Edge-Based Level Set Evolutions

As early as 1993, Caselles, *et al.* [49] proposed the following edge-based LSE:

$$\begin{cases} \phi_t = g(I)|\nabla\phi|\left(div\left(\frac{\nabla\phi}{|\nabla\phi|}\right) + \nu\right) \\ g(I) = \frac{1}{1+|\nabla G_\sigma * I|^2} \end{cases} \tag{2}$$

where $\phi$ denotes the LSF as before; $I$ represents the original image; $g(I)$ is an *edge function* with respect to the image gradient $\nabla I$; $div(\nabla\phi/|\nabla\phi|)$ is the mean curvature of ZLC [49]; $div$ is the divergence operator; $\nu > 0$ is a constant; $G_\sigma$ is a Gaussian kernel with the standard deviation $\sigma$; and $*$ is the convolution operator. In (2), $g(I)$ is the data term that attracts ZLCs toward the object boundaries, whereas the term $|\nabla\phi|div(\nabla\phi/|\nabla\phi|)$ is called regularization term that keeps the smoothness of ZLCs.

Later, Caselles, *et al.* [50] proposed a geodesic active contour (GAC) that is an extension of (2). Its level set formula is given below:

$$\phi_t = g(I)|\nabla\phi|\left(div\left(\frac{\nabla\phi}{|\nabla\phi|}\right) + \nu\right) + \nabla\phi \cdot \nabla g(I) \tag{3}$$

In comparison to model (2), GAC in (3) embraces an extra gradient term, namely, $\nabla\phi \cdot \nabla g(I)$ that is capable of pushing ZLCs toward the desired object boundaries, along which high gradient variations exist.

In traditional LSEs, LSF $\phi$ becomes irregular as it evolves, which finally leads to incorrect zero-level-sets. To obtain stable results, LSF should be periodically reinitialized to a signed distance function (SDF) during its evolution. However, the reinitialization step for LSF is problematic, both theoretically and practically [51]. To address this problem, Li, *et al.* [51] proposed a distance regularized level set evolution (DRLSE) that does not need to reinitialize LSF repeatedly and most importantly, it is able to maintain the LSF regular automatically. The evolution equation of DRLSE is given below:

$$\phi_t = \mu div\left(d_p(|\nabla\phi|)\nabla\phi\right) + \lambda\delta_\varepsilon(\phi)div\left(g\frac{\nabla\phi}{|\nabla\phi|}\right) + \alpha g\delta_\varepsilon(\phi) \tag{4}$$

where

$$d_p(s) = \begin{cases} \frac{\sin(2\pi s)}{2\pi s}, & if\ s \leq 1 \\ 1 - \frac{1}{s}, & if\ s \geq 1 \end{cases} \tag{5}$$

where $div\left(d_p(|\nabla\phi|)\nabla\phi\right)$ is called distance regularization term; $g$ is the *edge function* as before; $\delta_\varepsilon(\cdot)$ is the Dirac delta function; and $\mu, \lambda$, and $\alpha$ are free parameters.

### 2) Region-Based Level Set Evolutions

An early region-based LSE was proposed by Chan and Vese (CV) [52] in 2001, in which the region statistic (i.e., intensity mean) instead of the image gradient is used as the data term. Its corresponding level set formula is given as follows:

$$\phi_t = \delta_\varepsilon(\phi)\left[\mu div\left(\frac{\nabla\phi}{|\nabla\phi|}\right) - \nu - \lambda_1(I - c_{in})^2 + \lambda_2(I - c_{out})^2\right] \tag{6}$$

where $\mu$, $\nu$, $\lambda_1$, and $\lambda_2$ are free parameters. $c_{in}$ and $c_{out}$ are intensity means inside and outside ZLC, respectively.

From the perspective of statistics, CV in (6) can be viewed as a special case of the earlier motion equation proposed by Zhu and Yuille [53], in which the image intensity is assumed to be Gaussian:

$$\phi_t = \delta_\varepsilon(\phi)\left[\mu div\left(\frac{\nabla\phi}{|\nabla\phi|}\right) + \frac{(I-c_{out})^2}{2\sigma_{out}^2} - \frac{(I-c_{in})^2}{2\sigma_{in}^2} + log\frac{\sigma_{out}}{\sigma_{in}}\right] \quad (7)$$

where $\sigma_{in}^2$ and $\sigma_{out}^2$ are intensity variances inside and outside ZLC, respectively. $log$ is the natural logarithm. Its application for automatic man-made object extraction can also be found in [54].

Shortly after, Yezzi, *et al.* [55] also proposed a region-based curve evolution. For the case of two regions, the gradient flow of the curve evolution can be written as follows:

$$C_t = (u-v)(2*I-u-v)\vec{N} - \alpha\kappa\vec{N} \quad (8)$$

where $C$ denotes the ZLC; $u$ and $v$ are intensity means of the two regions, respectively; $\alpha$ is a free parameter; $\kappa = div(\nabla\phi/|\nabla\phi|)$ is the mean curvature as before; and $\vec{N}$ is the outward unit normal of the curve $C$. According to the *Eulerian formulation* in [56], the gradient flow (8) can be rewritten as:

$$\phi_t = [(u-v)(2*I-u-v) + \alpha\kappa]|\nabla\phi| \quad (9)$$

Based on the example in [57], we fix $\nu = 0$ and $\lambda_1 = \lambda_2 = 1$ in CV (6), at the same time, we replace $\delta_\varepsilon(\phi)$ by $|\nabla\phi|$, which is feasible as pointed out in [52]. Then CV can be rewritten as:

$$\phi_t = [(c_{in} - c_{out})(2*I - c_{in} - c_{out}) + \mu\kappa]|\nabla\phi| \quad (10)$$

As can be seen clearly, the model (8) proposed in [55] is a special case of CV (6).

More recently, based on GAC in (3), Zhang, *et al.* [58] proposed a simplified region-based LSE, in which the data term and regularization term were borrowed from [57] and [59], respectively. The formula is:

$$\phi_t = \nu\left[\frac{I - \frac{c_{in}+c_{out}}{2}}{max\left(\left|I - \frac{c_{in}+c_{out}}{2}\right|\right)}\right]|\nabla\phi| \quad (11)$$

where $\nu$ is a constant as in (3) and $max(\cdot)$ is the maximum operator. It is worth mentioning that the traditional regularization term is missing in (11). Instead, a Gaussian filter is introduced into the iterative process of LSE in a separate step to regularize the evolving ZLCs periodically. Additionally, due to the fact that $(I - c_{in} + I - c_{out}) = 2\left(I - \frac{c_{in}+c_{out}}{2}\right)$, it is clear that model (11) is also a special case of CV in (6).

### 3) Level Set Evolution in Remote Sensing

The earlier applications of LSE in the field of remote sensing can be found in [60-62]. Among them, LSE was employed for oil slick extraction from synthetic aperture radar (SAR) images [60], locating the discontinuity in sea surface temperature and soil moisture [61], and hyperspectral image classification [62].

LSEs have received increasing attention in the remote sensing community. In particular, CV model [52] becomes one of the most commonly used region-based LSEs for object extraction from remote sensing data, including man-made object extraction [54, 63-66], shoreline extraction [67], change detection [68, 69], building shadow extraction [70], and tree canopy reconstruction [71].

In addition to region-based LSEs, edge-based ones have also gained wide attention for object extraction in some studies. For example, GAC in (3) was employed to extract highways and vehicles from aerial images [72] and extract lakes and islands from Landsat images [73]. More recently, DRLSE (4) was utilized for automatic extraction of building rooftops from color aerial images [17].

### C. Limitations of State-of-the-Art Approaches

Although significant efforts have been made to develop practical methods for object extraction from remote sensing data, a couple of challenges still remain. To the best of our knowledge, the state-of-the-art methods (both non-LSEs and LSEs) are mainly developed for dealing with one specific type of object, rather than multiple kinds of objects.

General approaches take advantage of geometrical shapes, spectral information, or some other specific features such as shadow, which makes them effective in extracting some kinds of objects but may fail to extract others since different objects have their own intrinsic features.

On the other hand, existing LSEs are suffering from the following three limitations:

1) As seen in (4) and (6), there are many parameters that need to be tuned before they can be used in practice. Although the recommended values were provided in original studies, it is still difficult and labor-intensive for users to seek out the optimal parameters when they are employed in practical engineering applications. This will be further justified in our experiments.

2) In traditional LSEs, the LSF is commonly initialized as a SDF that needs to be reinitialized periodically in the iterative process. However, the computation for maintaining the LSF as a SDF is considerably expensive [59]. Although in some recent studies the reinitialization procedure was eliminated completely [51, 74], other problems arise. For instance, DRLSE in [51] has to spend more time on the computation of distance regularization term. While the reinitialization-free and region-based model in [74] can extract desired objects, it also extracts other spectral similarly ones in the same scene.

3) Essentially, the length regularization term, e.g., $\delta_\varepsilon(\phi)div(\nabla\phi/|\nabla\phi|)$, in traditional LSEs is a parabolic term and the time step should be subject to the Courant-Friedrichs-Lewy stability condition [51, 75], that is, it becomes impossible to use a large time step in the numerical scheme. In addition, it is computationally expensive [75]. All these factors result in slow convergence for traditional LSEs.

To sum up, the state-of-the-art methods still face challenges and the foregoing disadvantages might hamper their applications in some cases. Consequently, it is of great interest to develop new techniques that are more accurate and efficient, and most importantly, can handle multiple types of objects. For that purpose, we propose two fast LSEs in this study for man-made object extraction from optical remote sensing images.

## III. Proposed Level Set Evolutions

In this section, we first present the generalized form of LSE that may make its essence more intelligible. Then, we elaborate the proposed two fast LSEs: one is an edge-based and the other is a region-based. Finally, to demonstrate the advantages of the proposed LSEs, we compare them with each other and also with other state-of-the-art LSEs, by applying them to extract desired objects from a synthetic image.

### A. Generalized Form of Level Set Evolution

Based on the above recall of traditional LSEs, we can derive their generalized form as follows:

$$\phi_t = T_d + T_l + T_a \tag{12}$$

where $T_d$, $T_l$, and $T_a$ denote the data term, length regularization term, and area regularization term, respectively.

Data term $T_d$ is used to drive the propagating ZLCs toward the desired object boundaries. In edge-based LSEs, it is generally a function of the image gradient, e.g., the *edge function* $g(I)$ in (2), whereas in region-based LSEs it is commonly related to the region statistics, for example, the intensity means $c_{in}$ and $c_{out}$ on each side of ZLC in CV in (6). As for the length regularization terms $T_l$, e.g., $\delta_\varepsilon(\phi)div(\nabla\phi/|\nabla\phi|)$ in (6), it is generally related to the mean curvature and is used to keep the evolving ZLCs smooth and regular. However, the area regularization term $T_a$ is effective for removing small spurious objects [76], for example, $\nu\delta_\varepsilon(\phi)$ in (6). Note that $\delta_\varepsilon(\phi)$ can be sometimes replaced by $|\nabla\phi|$ [52, 74].

### B. Strategies for Improving Traditional Level Set Evolutions

As discussed before, traditional LSEs can be further improved both in terms of parameter tuning and computational efficiency. To this end, in this section we improve the traditional LSEs in three aspects, i.e., the LSF, the regularization term, and the time step.

*1) Level Set Function:* For computational efficiency, the following binary function (BF) [51, 75] is used as the LSF instead of the traditional SDF:

$$\phi(X, t = 0) = \begin{cases} 1, & X \in R \\ -1, otherwise \end{cases} \tag{13}$$

where $\phi(X, t = 0)$ is the initial LSF. $X$ denotes the coordinate of the pixel. $R$ is a region in the image domain. As presented in Fig. 1(a), each small rectangle corresponds to a pixel in image $I$. The purple rectangles represent the object region and the rest of the image is the background. The green line with black border denotes the initial ZLC. The matrix with values -1 and 1 is the initial LSF in (13), as shown in Fig. 1(b). The red zeros represent the zero-level-set and the red arrows indicate the evolution directions of ZLC. For more details of SDF, readers may refer to [77].

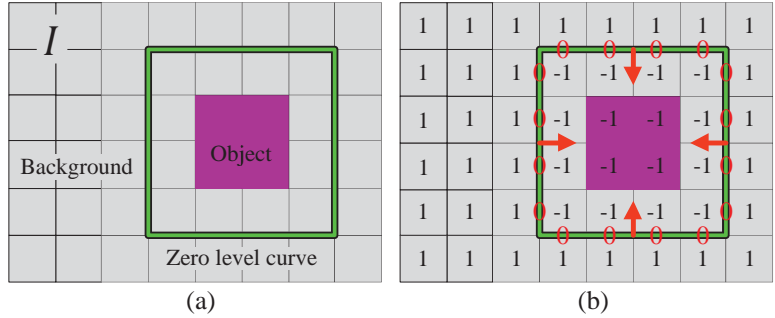

(a)  (b)

Fig. 1. Diagram of the LSF. (a) In image $I$, the purple pixels compose the object region and the rest of the image is the background. The green line with black border is the initial ZLC. (b) The matrix with values -1 and 1 represents the initial LSF. The red zeros represent the zero-level-set and the red arrows indicate the directions of the evolution.

*2) Regularization Term:* As stated earlier, there are generally two kinds of regularization terms (i.e., length and area regularization terms in (12)) in traditional LSEs in (3), (4), and (6). The former keeps ZLCs smooth, whereas the latter is utilized to filter out the small spurious patches. However, they usually cause expensive computation in practical applications [59]. Moreover, the involved free parameters often need laborious tuning. In fact, the length of ZLC is comparable with the area formed by itself [52]. Therefore, in traditional LSEs only the length regularization term is preserved, regardless of the area one. To improve the computational efficiency of LSE, in this context we replace traditional regularization terms by a Gaussian kernel as in [59]. It is not only able to smooth the moving ZLCs, but also capable of filtering out those small spurious patches. We can therefore eliminate the computationally costly regularization terms from traditional LSEs. Mathematically, it is viable to replace the traditional regularization term by a Gaussian kernel [27, 28].

*3) Time Step:* Due to the elimination of traditional length regularization term, we can employ a larger time step in the numerical scheme for the proposed LSEs, thereby expediting their evolutions substantially.

Based on the above three improvements of traditional LSEs, we can readily derive our fast level set formulations in the following, including the proposed edge- and region-based LSEs.

### C. Formulations of the Proposed Level Set Evolutions

In this section, we propose two fast level set formulations. In the following, we will describe the derivation of the two LSEs in detail and give an overview of their implementation.

*1) Formulation of the Proposed Edge-Based Level Set Evolution*

The proposed edge-based LSE in this study is established based on the classical ones in (2) and (3). Since the role of the traditional regularization term can be replaced by a Gaussian kernel, we can remove the term $div(\nabla\phi/|\nabla\phi|)$ from the edge-based LSE (2). Furthermore, since the constant $\nu$ in (2) and (3) is mainly used for speeding up the evolution of the ZLCs, it is often set to be zero for less parameter tuning [50]. In this way, we can obtain a new edge-base LSE as follows:

$$\begin{cases} \phi_t = g(I)|\nabla\phi| \\ g(I) = \frac{1}{1+|\nabla G_\sigma * I|^2} \end{cases} \quad (14)$$

where only two key free parameters need to be tuned before it can be employed in practice, namely the time step $\Delta t$ and the standard deviation $\sigma$ of the Gaussian kernel that is used to smooth the original image. Hereafter, the notation $E_{T_d}$ is employed to represent the edge-based data term $g(I)$ in (14). In comparison to GAC (3), the gradient term $\nabla\phi \cdot \nabla g(I)$ is not included in our LSE (14). As mentioned earlier, this gradient term is primarily developed for driving ZLCs to those boundaries along which there are gradient variations. However, the man-made objects in high spatial resolution remote sensing images often have clear boundaries, along which the gradient variations are rare. As a result, it is feasible to exclude this term from our LSE in (14). It should be noted, however, that the model (14) cannot be derived from DRLSE in (4) directly since (4) does not need the traditional reinitialization step of LSF during its evolution.

*2) Formulation of the Proposed Region-Based Level Set Evolution*

Similarly, we can derive a novel region-based LSE from CV [52] by removing the regularization term in

(6) and setting free parameters $\mu = \nu = 0$. Note that here we use the notation presented in [78] since it is more intuitive than that in (6) to some extent, namely

$$\phi_t = [-\lambda^+(I - c^+)^2 + \lambda^-(I - c^-)^2]\delta_\varepsilon(\phi) \tag{15}$$

where $c^+$ and $c^-$ are intensity averages on $\phi \geq 0$ and $\phi < 0$, respectively. $\lambda^+$ and $\lambda^-$ are corresponding coefficients. For less parameter tuning, we set $\lambda^+ = \lambda^- = 1$ similar to that in [52, 57]. In addition, we replace $\delta_\varepsilon(\phi)$ by $|\nabla\phi|$ since it is computationally more efficient. Therefore, the model (15) can be rewritten as:

$$\phi_t = (c^+ + c^-)(2I - c^+ - c^-)|\nabla\phi| \tag{16}$$

To obtain more stable numerical results, we normalize the data term $(c^+ - c^-)(2I - c^+ - c^-)$ as in [58] and thus (16) can be further written as:

$$\phi_t = \left[ \frac{(c^+ - c^-)(2I - c^+ - c^-)}{max(|(c^+ - c^-)(2I - c^+ - c^-)|)} \right] |\nabla\phi| \tag{17}$$

Finally, we obtain a new region-based LSE, in which only one parameter (i.e., time step $\Delta t$) need to be given in advance. Most importantly, a large time step can be used in the numerical scheme, thus expediting the convergence of (17) considerably. Hereafter, we employ the notation $R_{T_d}$ to represent the region-based data term $\frac{(c^+ - c^-)(2I - c^+ - c^-)}{max(|(c^+ - c^-)(2I - c^+ - c^-)|)}$ in (17). In the next section, we will describe the implementation of the two types of LSEs in detail.

*D. Implementation of the Proposed Level Set Evolutions*

As presented in Table I, the implementation of the proposed LSEs mainly consists of six steps.

TABLE I
PSEUDO-CODE OF THE PROPOSED LSEs

| | |
|---|---|
| **Input:** | Image $I$ |
| | Scale parameters $\sigma_1$ and $\sigma_2$ for the edge-based LSE (14) or |
| | Scale parameter $\sigma$ for the region-based LSE (17) |
| | Initial ZLC |
| | Time step $\Delta t$ |
| **Output:** | Binary result of the desired object |

1. Initialize the LSF $\phi$ as (13), including its position and signs on each side of ZLCs
2. Level set evolution
   2.1. Compute the data term and $|\nabla\phi|$
      $E_{T_d} = g(I)$
      $R_{T_d} = \dfrac{(c^+ - c^-)(2I - c^+ - c^-)}{max(|(c^+ - c^-)(2I - c^+ - c^-)|)}$
   2.2. Update the LSF $\phi$
      $\phi_{i,j}^{n+1} = \phi_{i,j}^n + \Delta t R(\phi_{i,j}^n)$
3. Reinitialize the updated LSF $\phi$ if necessary
   $\phi = 1$, if $\phi > 0$; otherwise, $\phi = -1$
4. Regularize the updated LSF $\phi$ using a Gaussian kernel
   $\phi = G_\sigma * \phi$
5. If it has not yet converged, goto 2
6. Return the final LSF $\phi$, i.e., the binary result of the desired object

*Step 1:* The LSF $\phi$ in proposed LSEs is initialized as BF in (13), which is not only easy to use, but computationally efficient. More precisely, the initialization of LSF includes two steps: 1) finding the appropriate positions for the initial ZLCs, and 2) determining the signs of the LSF on each side of ZLCs (see Fig. 1).

*Step 2:* Once the initial LSF is determined, ZLCs can be evolved by using the proposed LSE in (14) or (17). In this step, we first compute the data term (i.e., $E_{T_d}$ or $R_{T_d}$) and $|\nabla\phi|$. Then, we update the LSF using the following finite difference scheme [51, 52]:

$$\phi_{i,j}^{n+1} = \phi_{i,j}^n + \Delta t R(\phi_{i,j}^n) \tag{18}$$

where $(i, j)$ is the spatial position; $n$ is the iteration number; $\Delta t$ is the time step; and $R(\cdot)$ denotes the right hand side of the proposed LSEs in (14) and (17). It is particularly worth mentioning that the computation of the data term in the proposed edge-based LSE in (14) requires original images to be convolved with a Gaussian kernel, whereas it is unnecessary for the proposed region-based LSE in (17).

*Step 3:* For the proposed region-based LSE (17), this step is optional in some practical applications like in [52]. When only a small number of objects need to be extracted from a complex scene, in which there are many undesired objects that generally have similar intensities to the desired ones, we need to reinitialize the LSF periodically to ensure that $|\nabla\phi| \neq 0$ around ZLCs while $|\nabla\phi| = 0$ elsewhere (see Fig. 1). In this way, the LSE is only effective around the desired objects, thereby avoiding extracting other undesired objects. Specifically, $\phi$ is reinitialized as BF in (13), i.e., let $\phi = 1$, if $\phi > 0$; otherwise, let $\phi = -1$. Sometimes, when multiple objects need to be extracted simultaneously we do not need to reinitialize the LSF. In this case, there is $|\nabla\phi| \neq 0$ across the entire image gradually, thus making the LSE effective for every pixel. It is worth mentioning however that this situation just holds for images with fewer noises. To ensure that the favorable results can be obtained in our experiments, we reinitialize the LSF periodically since the desired objects are generally surrounded by heavy noises in complex scenes.

For the proposed edge-based LSE (14), on the other hand, the LSF should be reinitialized periodically; otherwise, it will become nonnegative according to (18) and further the ZLC will vanish since the date term $g(I)$ is always greater than or equal to 1.

*Step 4:* Since the typically used mean curvature-based regularization term is eliminated from the proposed LSEs, we employ a Gaussian kernel instead to maintain the moving ZLC smooth and regular. That is, we convolve the updated LSF periodically with a Gaussian kernel. Certainly, the scale parameter $\sigma$ of the Gaussian kernel can be tuned accordingly for different applications that will be discussed in the experiments. Traditionally, in order to remove those small undesired patches from final results, morphological operations are often introduced into this step, as was done in [17]. However, they are unnecessary for the proposed LSEs, since the Gaussian kernel is able to keep ZLC smooth while filtering out those small undesired objects.

*Step 5:* Determine whether the convergence is reached; otherwise, go back to Step 2. In fact, once ZLC stops at the object boundary, the LSF will not be updated any more. Therefore, the iteration can be terminated if the LSF stop varying.

*Step 6*: Output the final LSF that is essentially a binary version of the extracted objects.

Finally, it is worth noting that there are two Gaussian kernels used to implement the proposed edge-based LSE (14). One is used for the denoising procedure in Step 2.1 in Table I to smooth the original image so that more appropriate data term $g(I)$ can be obtained. The other is employed in Step 4 to ensure the smoothness of the updated LSF. Thus, the two Gaussian kernels play different roles. The proposed region-based LSE (17), by contrast, does not need the denoising. However, it requires a Gaussian kernel in Step 4 to keep the evolving LSF regular. Hereafter, we refer to the scale parameters of the two Gaussian kernels used in (14) as $\sigma_1$ and $\sigma_2$, respectively, whereas the scale parameter used in (17) as $\sigma$.

In the next two sections, we will further describe the details on the implementation of the proposed LSEs by giving some simple examples. Meanwhile, we compare the two proposed LSEs with each other (Section III-E) and also compare them with other state-of-the-art LSEs (Section III-F).

### E. Comparison Between the Proposed LSEs

In this study, two LSEs are presented: one is the edge-based LSE (14) and the other is the region-based LSE (17). In this section, we apply them to extract objects from synthetic images and compare them with each other in the following three aspects.

### 1) Sensitivity to the Initial Position of Zero Level Curve:

It is evident that the edge-based date term $E_{T_d}$ in (14) is always positive in the image domain (except for the boundaries where it is approximately zero), whereas the region-based data term $R_{T_d}$ in (17) can be positive or negative on both sides of ZLCs. This property has a direct effect on the propagation direction of ZLCs. As shown in Fig. 2, it makes the edge-based LSE (14) sensitive to the initial position of ZLCs.

In this example, we fix the LSF $\phi$ to be negative inside ZLC and positive outside. We first put the initial ZLC outside the object boundary, as shown in the left panel of Fig. 2(a). Finally, both the edge- and region-based LSEs extract the desired object accurately, as presented in the middle and right panels of Fig. 2(a), respectively. However, when the initial ZLC is put at other positions such as the case demonstrated in the left panel of Fig. 2(b), that is, the initial ZLC intersects the object boundary, which is common in practice, the edge-based LSE finally fails to detect the desired object. This is because the date term $g(I)$ is always positive in both object and background areas, and thus, ZLC evolves inward simultaneously, as indicated by the red arrows in the left panel of Fig. 2(b). Finally, ZLC shrinks and vanishes. The intermediate iteration and final result are shown in the middle and right panels of Fig. 2(b).

With the same initialization of ZLC, the region-based LSE, by contrast, has better performance since its data term has opposite signs on each side of ZLC (see Fig. 3 for details). As indicated by the red arrows in the left panel of Fig. 2(c), the evolution directions of ZLC are opposite and it is finally attracted to desired boundaries accurately, see the middle and right panels of Fig. 2(c), respectively.

This example indicates that the initial ZLC for the proposed edge-based LSE (14) should be strictly put at either the object region or the background area. By contrast, the initial position of ZLC for the proposed region-based LSE (17) is more flexible and theoretically, it can be put at anywhere in the image [52].

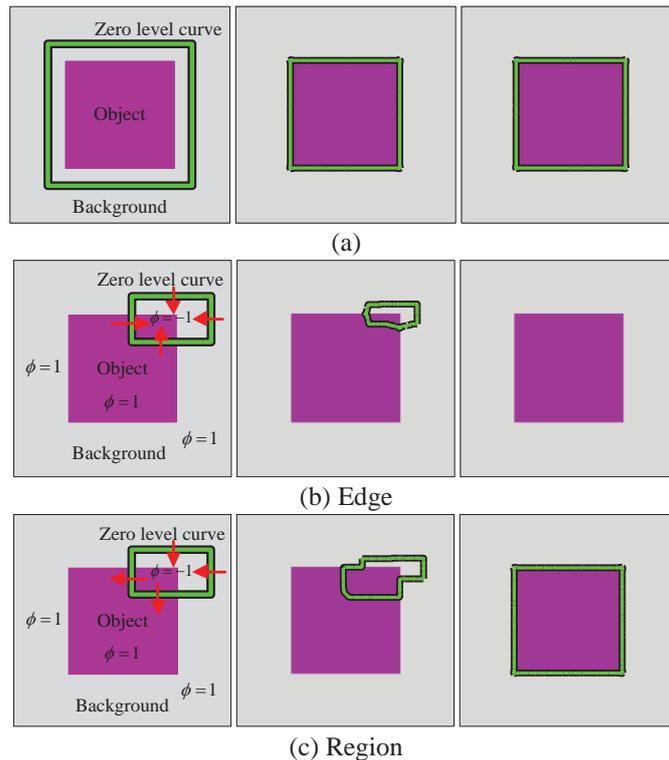

Fig. 2. Results of the proposed LSEs with different initial positions of ZLC, where LSF $\phi$ is set negative inside ZLC and positive outside. (a) Results of the proposed LSEs (14) and (17) with initial ZLC outside the object boundary. Left panel: the initial ZLC. Middle and Right panels: results of the proposed edge- and region-based LSEs. (b) and (c) Results of the edge- and region-based LSEs with initial ZLC crossing the object boundary. Left panel: the initial ZLC with evolution directions. Middle and Right panels: intermediate and final results of the proposed LSEs.

*2) Sensitivity to the Sign of LSF:* In addition to the sign of the date term, the sign of LSF is also a key determinant of LSE. We found that the proposed edge-based LSE (14) is sensitive to the sign of LSF, that is, the sign of LSF on each side of ZLC should be strictly set before it begins to evolve, which is time-consuming and unfavorable in practical applications.

For better understanding, we present an example in Fig. 3. Here, we use the same image $I$ as in Fig. 2. Both the object and background areas are assumed to be homogeneous and their intensities are subject to the constraint: $I_O < I_B$, where $I_O$ and $I_B$ are object and background intensity, as can be seen in Fig. 3. Without loss of generality, we put the initial ZLC at the background region. For the data term $E_{T_d}$ in (14), there is always $E_{T_d} > 0$ in both the object and background regions while $E_{T_d} \approx 0$ on the object boundaries. To push ZLC toward object boundaries, we should set LSF $\phi$ negative inside it and positive outside, as shown in Fig. 3(a). In this case, the negative LSF $\phi$ will become positive according to the iteration process (18). That means ZLC moves inward, as indicated by the red arrows. When it arrives at object boundaries, LSF $\phi$ will not vary again because $E_{T_d} \approx 0$. Conversely, if LSF $\phi$ is set negative outside it and positive inside (Fig. 3(b)), it will expand outward and finally fails to extract the desired object. Therefore, the proposed edge-based LSE (14) is sensitive to the signs of LSF.

By contrast, the proposed region-based LSE (17) is robust to the signs of LSF. As demonstrated in Fig. 3(c) and (d), no matter what kind of LSF is used (i.e., negative inside or outside ZLC), it always extracts the desired object. As presented in Fig. 3(c), when the LSF is negative inside ZLC and positive outside, in the object region, there is $(c^+ - c^-) > 0$, $(I - c^+) < 0$, $(I - c^-) < 0$, and $(2I - c^+ - c^-) < 0$, and thus, the data term $R_{T_d} < 0$; in the background area, there is $(c^+ - c^-) > 0$, $(I - c^+) = 0$, $(I - c^-) > 0$, and $(2I - c^+ - c^-) > 0$, and finally the data term $R_{T_d} > 0$. Since the initial position of ZLC is within the background region, the negative LSF $\phi$ will become positive based on the iteration equation (18). That is, ZLC moves inward. Similarly, when the LSF is positive inside and negative outside ZLC, as shown in Fig. 3(d), the region-based LSE (17) can also extract the desired object satisfactorily.

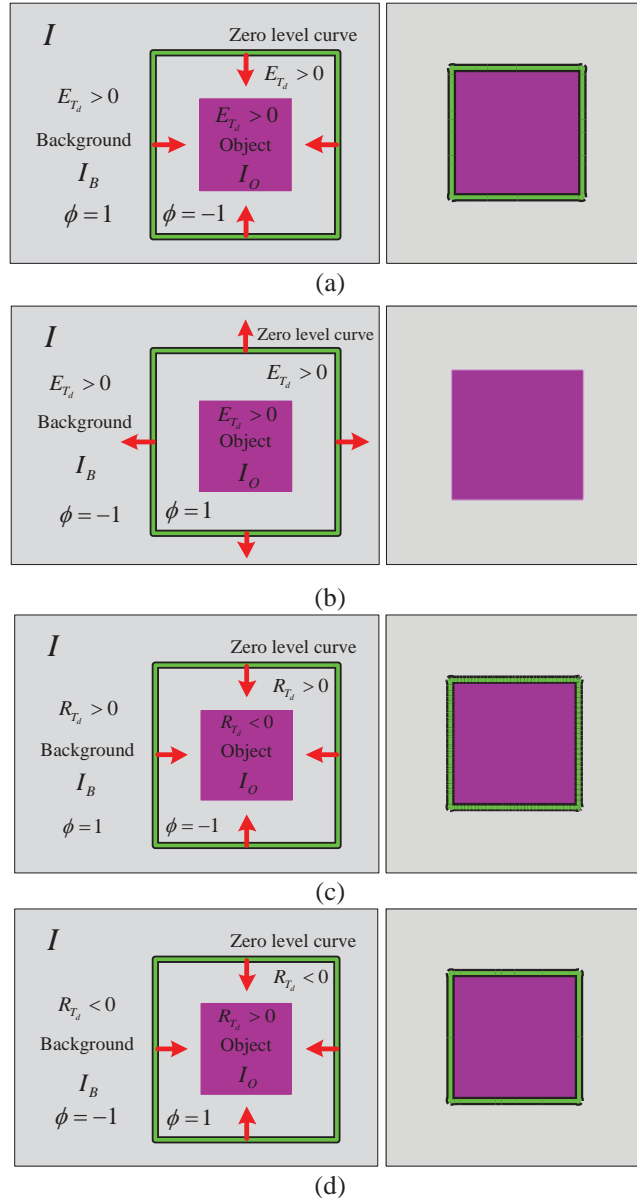

Fig. 3. Results of the proposed edge- and region-based LSEs with different signs of LSF $\phi$ on each side of ZLC. $I_O$ denotes the object intensity, whereas $I_B$ is the background intensity. $E_{T_d}$ and $R_{T_d}$ are the data terms of the edge- and region-based LSEs. The green line with the black border represents ZLC. The red arrows indicate the evolution directions. (a) and (b) Results of the edge-based LSE (14) with LSF negative inside and outside ZLC and the other way around, respectively. (c) and (d) Results of the region-based LSE (17) with LSF negative inside and outside ZLC and the other way around, respectively.

*3) Sensitivity to Noise:* As stated earlier, two Gaussian kernels are used in the implementation of the proposed edge-based LSE. One is used to smooth the original image (i.e., eliminating small spurious artifacts), whereas the other is employed to keep the LSF regular. In comparison, the region-based LSE is much more robust to image noise and it therefore does not need the denoising procedure.

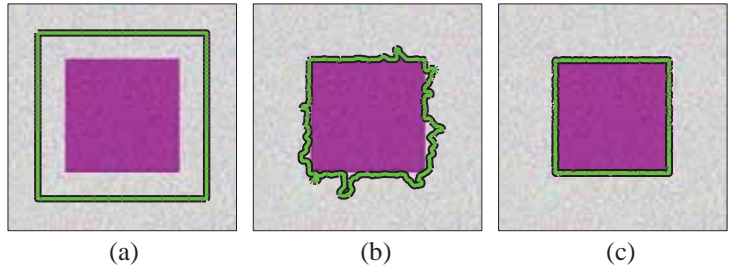

(a)         (b)         (c)

Fig. 4. Object extraction from the noisy image using the proposed LSEs. (a) Original image with Gaussian white noise of zero mean and standard deviation $\sigma = 0.2$. The green line with black border is the initial ZLC. (b) Result of the proposed edge-based LSE (14). (c) Result of the proposed region-based LSE (17).

As presented in Fig. 4(a), we add a Gaussian white noise of zero mean and standard deviation $\sigma = 0.2$ to the original image used in Fig. 2. As shown in Fig. 4(b), the edge-based LSE (14) is sensitive to the image noise. Due to the noise, the data term $g(I)$ becomes pointless. Checking Fig. 4(c), despite the image noise, the proposed region-based LSE extracts the desired object accurately.

Based on the above analysis, it is clear that the region-based LSE (17) is more robust than the edge-based LSE (14) in terms of the initialization of ZLC, the sign of LSF, and the image noise on synthetic images. However, that does not follow that the region-based LSE (17) will definitely perform better than the edge-based LSE (14) in all real cases. In the following experiments of man-made object extraction from real remote sensing images, the strengths and weaknesses of the two LSEs will be further investigated.

### F. Comparison with Other State-of-the-Art Approaches

In this section, to demonstrate the advantages of the proposed two LSEs, we compare them with other state-of-the-art approaches in several aspects. First, we compare the proposed edge-based LSE (14) with the popular DRLSE in (4) since they both belong to the edge-based LSE. Then we discuss the similarity and distinction between the proposed region-based LSE (17) and the model (11) proposed in [58].

Compared with DRLSE, the main advantage of the proposed edge-based LSE (14) is threefold. First, it is computationally more efficient since a simpler LSF (i.e., BF (13)) is employed in the numerical scheme. Second, due to the substitution of a Gaussian kernel for the traditional regularization terms, a larger time step can be used for (14) in the numerical scheme, thus making its convergence faster (see Fig. 12 (d)). Last but not least, the edge-based LSE (14) will be more operational than DRLSE in practical applications since it needs much less parameter tuning, as presented in Table III.

The similarity between the LSEs (11) and (17) is twofold. First, both the data terms in the two LSEs are adapted from the CV [52], and thus, they generally have similar accuracy for object extraction, as shown in Fig. 12(a)-(c). Second, to avoid numerical instability, they both normalize the original data term in their formulations.

In addition, it is particularly worth mentioning that there are two essential distinctions between the LSEs (11) and (17). First, the data term in the proposed LSE (17) has one more term than that in (11), i.e., $(c^+ - c^-)$, which makes (17) more robust to the signs of LSF, as presented in Fig. 3(c) and (d). By contrast, the model (11) is sensitive to the signs of LSF, as shown in Fig. 5. Similar to the examples in Fig. 3(c) and (d), the initial ZLC is put at the background region. In this case, according to the data term in (11), there is $(I - c_{in}) < 0$, $(I - c_{out}) < 0$, and further $(2I - c_{in} - c_{out}) < 0$ in the object region, whereas in the background region, there is $(I - c_{in}) > 0$, $(I - c_{out}) = 0$, and finally $(2I - c_{in} - c_{out}) > 0$. To drive ZLC toward the desired object boundary, we should strictly set the LSF negative inside and positive outside ZLC, respectively, as presented in Fig. 5(a). Otherwise, it would fail to extract the desired object, as shown in Fig. 5(b). Therefore, the model (11) proposed in [58] is sensitive to the signs of LSF.

The second distinction between LSEs (11) and (17) is that they are built on different foundations. Our LSE (17) is derived from the region-based CV [52] and [78] directly, whereas the model (11) stems from the edge-based GAC [50] and thus there is one more free parameter (i.e., $\nu$) in it.

For a further demonstration of the advantages of the proposed LSEs (14) and (17), in the next section we compare them with the state-of-the-art LSEs in extracting multiple kinds of man-made objects from different remote sensing images qualitatively and quantitatively.

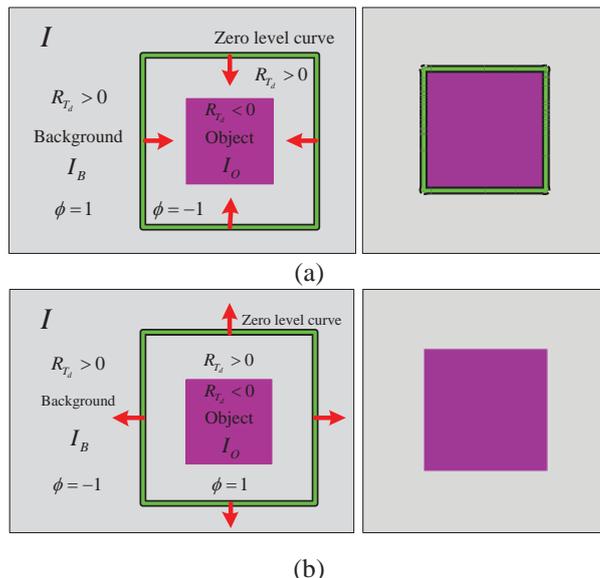

Fig. 5. Results of the region-based LSE (11) proposed in [58] with different signs of LSF $\phi$ on each side of ZLC. $I_O$ denotes the object intensity, whereas $I_B$ is the background intensity. $R_{T_d}$ is the data term in (11). The green line with the black border represents ZLC. The red arrows indicate the evolution directions. (a) and (b) Results of the region-based LSE (11) with LSF negative inside and outside ZLC, respectively.

## IV. EXPERIMENTAL RESULTS

### A. Dataset Description and Experiment Setup

In this study, five images obtained from Google Maps are used for the test. According to their order of appearance in the following, they are named building_1, building_2, airport_1, airport_2, and road, respectively. Both building_1 and _2 are situated in Hilo, Hawaii, USA, whereas the road is located at Germantown, Milwaukee, USA. As for the airport data, they are the George Bush International Airport and Hartsfield Jackson Atlanta International Airport, USA, respectively. They all contain three bands (i.e., R, G, and B). The spatial resolution and size of each image are given in Table II. The experiments are run under MATLAB R2013a 64 b on Window 7 OS with a Pro of Intel(R) Core(TM) i7-3770 CPU @ 3.40GHz, 16GB RAM. We intend to make our MATLAB code publically available at 'http://www.lsgi.polyu.edu.hk/academic_staff/John.Shi/index.htm'.

TABLE II
DATASET DESCRIPTION

| Images | Spatial resolution (meter) | Size (pixel×pixel) | Location |
|---|---|---|---|
| Building_1 | 0.3 | 431×507 | Hilo, Hawaii, USA |
| Building_2 | 0.25 | 553×461 | Hilo, Hawaii, USA |
| Airport_1 | 2.0 | 1792×1536 | Houston, Texas, USA |
| Airport_2 | 1.0 | 2246×2234 | Atlanta, GA, USA |
| Road | 0.7 | 731×1507 | Germantown, Milwaukee, USA |

To validate the advantages of the proposed LSEs, we compare them with other two state-of-the-art LSEs, i.e., the DRLSE [51] and CV [52] qualitatively and quantitatively. For qualitative evaluation, we apply them to extract three types of man-made objects, i.e., building roofs, road networks, and airport runways from the images mentioned above. The aim is to demonstrate the abilities of the proposed LSEs according to the following four aspects:

1) automatic topology changes of ZLC, namely, the splitting and merging;
2) dealing with isolated and extraneous small objects;
3) handling the occlusions or discontinuities caused by nearby objects;
4) detecting desired objects with various complicated geometric shapes from complex scenes.

For fair comparison, in each experiment we initialize ZLCs for each LSE at the same positions and choose the best performance for each method via trial-and-error test. The free parameters used for them are given in Table III.

For quantitative evaluation, we employ three commonly used indices, i.e., $Completeness = P_m/P_g$,

$Correctness = P_m/P_e$, and $Quality = P_m/(P_e + P_{um})$, in which $P_m$ denotes the total pixels of the extracted object that are matched with the ground truth data, $P_g$ is the total pixels of the ground truth data, $P_e$ is the total pixels of the extracted object, and $P_{um}$ is the total pixels of the ground truth data that are unmatched with the extracted object. Ground truths are acquired by digitizing the original images manually. Additionally, the CPU time, another critical metric, is recorded for each method and used for the evaluation of their computational efficiency.

TABLE III
FREE PARAMETERS USED FOR EACH LSEs AND IMPORTANT PARAMETERS ARE HIGHLIGHTED (UNIT FOR $\Delta t$: SECOND )

| Experiments | Proposed LSE (17) | Proposed LSE (14) | DRLSE | CV |
|---|---|---|---|---|
| Building_1 in Fig. 6 | $\Delta t = 15$, $\sigma = 1$. and TS 9×9 in (b) $\Delta t = 15$, $\sigma = 2$. and TS 9×9 in (c) $\Delta t = 15$, $\sigma = 3$. and TS 9×9 in (d) | $\Delta t = 15$, $\sigma_1 = 1$, $\sigma_2 = 1.5$, and TS 9×9 in (e) $\Delta t = 15$, $\sigma_1 = 3$, $\sigma_2 = 1.5$, and TS 9×9 in (f) $\Delta t = 15$, $\sigma_1 = 5$, $\sigma_2 = 1.5$, and TS 9×9 in (g) | $\Delta t = 5$, $\lambda = 5$, $\mu = 0.04$, $\sigma = 3$, $c_0 = 2$, $\alpha = 1$, and TS 15×15 in (h) $\Delta t = 5$, $\lambda = 5$, $\mu = 0.04$, $\sigma = 3$, $c_0 = 2$, $\alpha = 2$, and TS 15×15 in (i) $\Delta t = 5$, $\lambda = 5$, $\mu = 0.04$, $\sigma = 3$, $c_0 = 2$, $\alpha = 3$, and TS 15×15 in (j) | $\Delta t = 0.8$, $\lambda_1 = \lambda_2 = 1$, $\nu = 0$, and $\mu = 0.1$ in (k) $\Delta t = 0.8$, $\lambda_1 = \lambda_2 = 1$, $\nu = 0$, and $\mu = 0.3$ in (l) $\Delta t = 0.8$, $\lambda_1 = \lambda_2 = 1$, $\nu = 0$, and $\mu = 0.5$ in (m) |
| Building_2 in Fig. 7 | $\Delta t = 15$, $\sigma = 4$. and TS 9×9 in (b)-(d) | $\Delta t = 18$, $\sigma_1 = 3$, $\sigma_2 = 1.5$, and TS 9×9 in (e)-(g) | $\Delta t = 5$, $\lambda = 5$, $\mu = 0.04$, $c_0 = 2$, $\sigma = 5, \alpha = 2$, and TS 13×13 in (h) $\Delta t = 5$, $\lambda = 5$, $\mu = 0.04$, $c_0 = 2$, $\sigma = 4, \alpha = 2.5$, and TS 13×13 in (i) $\Delta t = 5$, $\lambda = 5$, $\mu = 0.04$, $c_0 = 2$, $\sigma = 4, \alpha = 3$, and TS 13×13 in (j) | $\Delta t = 0.8$, $\lambda_1 = \lambda_2 = 1$, $\nu = 0$, and $\mu = 0.1$ in (k) $\Delta t = 0.8$, $\lambda_1 = \lambda_2 = 1$, $\nu = 0$, and $\mu = 0.3$ in (l) $\Delta t = 0.8$, $\lambda_1 = \lambda_2 = 1$, $\nu = 0$, and $\mu = 0.5$ in (m) |
| Airport_1 in Fig. 8 | $\Delta t = 18$, $\sigma = 2$, and TS 15×15 in (b) | $\Delta t = 16$, $\sigma_1 = 1$, $\sigma_2 = 1$, and TS 9×9 in (c) | $\Delta t = 5$, $\lambda = 5$, $\mu = 0.04$, $c_0 = 2$, $\sigma = 1.5, \alpha = 3.5$, and TS 7×7 in (d) | $\Delta t = 1.0$, $\lambda_1 = \lambda_2 = 1$, $\nu = 0$, and $\mu = 0.2$ in (e) |
| Airport_2 in Fig. 9 | $\Delta t = 18$, $\sigma = 3$. and TS 9×9 in (b) | $\Delta t = 18$, $\sigma_1 = 1$, $\sigma_2 = 1$, and TS 9×9 in (c) | $\Delta t = 5$, $\lambda = 5$, $\mu = 0.04$, $c_0 = 2$, $\sigma = 2, \alpha = 3$, and TS 9×9 in (d) | $\Delta t = 1.0$, $\lambda_1 = \lambda_2 = 1$, $\nu = 0$, and $\mu = 0.2$ in (e) |
| Road in Fig. 10 | $\Delta t = 18$, $\sigma = 3$. and TS 9×9 in (b) | $\Delta t = 18$, $\sigma_1 = 1.7$, $\sigma_2 = 1$, and TS 9×9 in (c) | $\Delta t = 5$, $\lambda = 5$, $\mu = 0.04$, $c_0 = 2$, $\sigma = 1.5, \alpha = 3.8$, and TS 7×7 in (d) | $\Delta t = 1.0$, $\lambda_1 = \lambda_2 = 1$, $\nu = 0$, and $\mu = 0.2$ in (e) |

### B. Qualitative Evaluation

#### 1) Experiment 1: Building Roof Extraction

In this section, we first take advantage of the splitting of ZLCs in LSEs to extract multiple objects simultaneously. Then, we employ their merging to extract a single building roof. The advantages of the merging of ZLCs will be further verified in the experiments of airport runway extraction and road network extraction.

Fig. 6 presents the results of LSEs (17), (14), DRLSE, and CV for simultaneous extraction of multiple building roofs by splitting the initial ZLC. In this experiment, region-based LSEs, i.e., the proposed (17) and CV clearly outperform edge-based ones, i.e., the proposed (14) and DRLSE. More specifically, the performance of LSE (17) is comparable to that of CV, whereas LSE (14) excels DRLSE clearly.

The original image (i.e., building_1) with initial ZLC and the corresponding ground truth are shown in the left and right panels of Fig. 6(a), respectively. Note that the white small object indicated by the red arrow is a building roof, whereas the one indicated by the yellow arrow is a truck. In each pair of Fig. 6(b)-(m), the left panel demonstrates the final ZLCs and the corresponding binary results are shown on the right. Specifically, Fig. 6(b)-(d) presents the results of LSE (17) with different scale parameters, i.e., $\sigma = 1, 2$, and 3, for the Gaussian kernel, respectively. As shown clearly, the initial ZLC is automatically split into independent parts and each part corresponds to a single building roof; however, other undesired small objects are also extracted. From left to right panels in Fig. 6(b)-(d), with the increase of the scale parameter $\sigma$, those smaller extraneous objects are removed effectively and only the desired objects are extracted finally. Thus, a relatively larger value of $\sigma$ is often advised in practical applications to ensure better anti-noise performance. Despite that, the truck is still erroneously recognized as a building roof, as indicated by the yellow arrow in the left panel of Fig. 6(a). This is due to the fact that they have similar spectral characteristic. In this case, to obtain favorable results, further post-processing is needed. In addition, the scale parameter $\sigma$ cannot be too large, otherwise LSE (17) may fail to extract the small building roofs such as the one indicated by the red arrow in the left panel of Fig. 6(a).

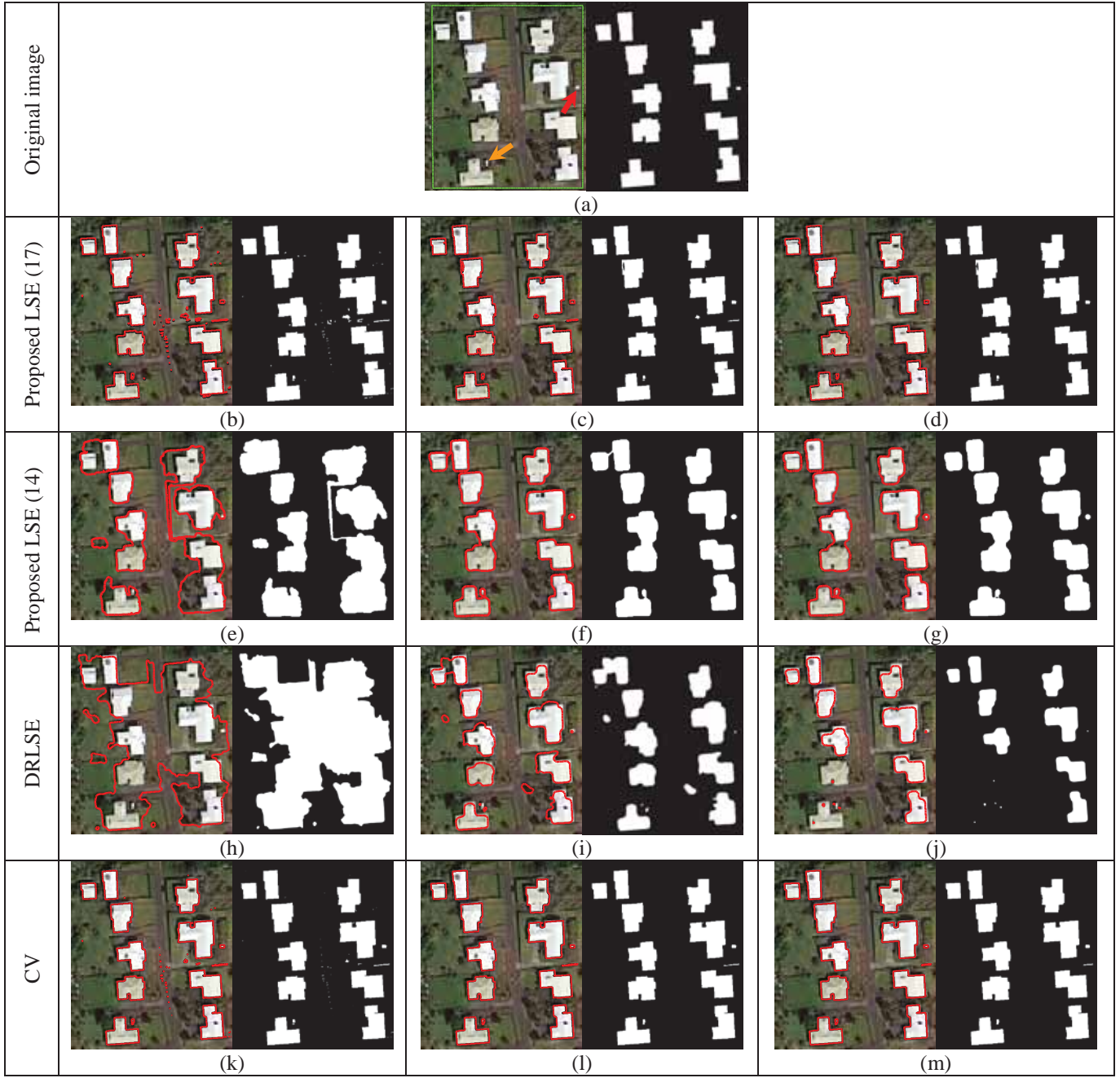

Fig. 6. Results of LSEs (17), (14), DRLSE, and CV for building roof extraction. (a) Left panel: the original image (i.e., building_1) with the initial ZLC. Right panel: the ground truth. The white small object indicated by the red arrow is a building roof, whereas the one indicated by the yellow arrow is a truck. (b)-(d) Left panel in each pair: final ZLCs of LSE (17) with the scale parameter $\sigma = 1, 2,$ and 3, respectively. Right panel in each pair: corresponding binary results. (e)-(g) Results of the proposed edge-based LSE (14) with the scale parameter $\sigma_1 = 1, 3,$ and 5, respectively. (h)-(j) Results of DRLSE with $\alpha = 1, 2,$ and 3, respectively. (k)-(m) Results of CV with $\mu = 0.1, 0.3,$ and 0.5, respectively. See Table III for detailed parameter tuning.

The results of the proposed edge-based LSE (14) are shown in Fig. 6(e)-(g). From left to right panels, the original image is convolved by a Gaussian kernel with the scale parameters $\sigma_1 = 1, 3,$ and 5, respectively.

With the increase of the scale parameter, the original image becomes smoother, suggesting that the results are more accurate. Nevertheless, two separated building roofs are extracted as a single one due to the failure of the splitting of ZLC, as presented in Fig. 6(g). In addition, the small truck is also wrongly recognized as a building roof.

Fig. 6(h)-(j) presents the results of DRLSE with $\alpha = 1$, 2, and 3, respectively. It is clear that with the increase of the parameter $\alpha$, more accurate boundaries can be extracted. However, a larger value of $\alpha$ may lead to boundary leakage. As shown in Fig. 6(j), two darker building roofs are finally missing. In comparison, a relatively small value of $\alpha$ can guarantee the extraction of all objects, as shown in Fig. 6(i). However, it also extracts other objects.

Finally, the results of CV with $\mu = 0.1$, 0.3, and 0.5 are presented in Fig. 6(k)-(m), respectively. As we can see, the smaller extraneous objects are filtered out with the increase of the parameter $\mu$. The optimal result is given in Fig. 6(m).

Fig. 7 presents the results of all the LSEs for a single building roof extraction. The goal of this experiment is to verify that LSEs can merge separate ZLCs automatically during their evolutions. Experiments show that the proposed LSEs (14) and (17) are able to extract the desired single building roof with discontinuities, whereas DRLSE and CV fail.

The original image (i.e., building_2) and the ground truth are shown in the left and right panels of Fig. 7(a), respectively. As can be seen, there are intensity variations on the building roof, which leads to discontinuities and thus poses a great challenge for traditional LSEs. Unlike in Fig. 6, the initial ZLCs are put at the object region. The intermediate processes of LSE (17) at iterations 15, 18, and 82 are presented on the left panel of Fig. 7(b)-(d), respectively. The corresponding binary results are given on the right panel. We found that the two separate ZLCs begin to merge with each other at iteration 15, as shown in Fig. 7(b). Thereafter, a new ZLC is generated at iteration 18 and it continues evolving in the following iterations until it converges. Although two ZLCs are initialized in the beginning, there is always only one LSF (i.e., the matrix with values 1 and -1 in Fig. 1) changes from iteration to iteration. In this experiment it was found that the proposed LSE (17) cannot only filter out those small foreign objects present on the building roof automatically, but also handle the discontinuities effectively. The only disadvantage of LSE (17) is that it also extracts some nearby undesired objects.

The results of the edge-based LSE (14) are illustrated in Fig. 7(e)-(g). The ZLCs and the corresponding binary results at iterations 18, 24, and 148 are presented from left to right panels, respectively. It can be observed that the merging of ZLCs commences at the 18th iteration. After that, a new ZLC emerges and it goes on to evolve until it stops at the desired boundary. Similar to (17), it also filters out those small extraneous objects automatically while passing through the discontinuities successfully. Due to the convolution of the original image with a Gaussian kernel, the edges extracted by the edge-based LSE (14) are smoother than those extracted by the region-based LSE (17).

The results of DRLSE for the single building roof extraction are shown in Fig. 7(h)-(j). Different from the previous experiments, here the results for specific iterations of the merging of ZLCs are not provided. Instead, we just illustrate the results of DRLSE for different parameters used (see Table III). As presented in the results, although the original image in DRLSE is also convolved with a Gaussian kernel as in LSE (14), ZLC cannot pass through the discontinuities and thus it fails to extract the whole object. Using larger values of $\sigma$ and $\alpha$, it is sometimes able to extract more objects, as shown in Fig. 7(i). However, that generally also causes serious edge leakages, see Fig. 7(j). To obtain the accurate object in some practical applications, we therefore suggest using relatively small values of $\sigma$ and $\alpha$.

The final results of CV with $\mu = 0.1$, 0.3, and 0.5 are shown in Fig. 7(k)-(m), respectively. In this experiment, the small spurious objects can be filtered out automatically by using a larger value of $\mu$. However, no matter which value of $\mu$ we use, CV cannot deal with the discontinuities and finally fails to extract the complete object. Actually, the discontinuities divide the seemingly complete building roof into several separate parts. In addition to that, no smoothing filter is used to convolve the original image for CV. All these factors cause the failure of CV.

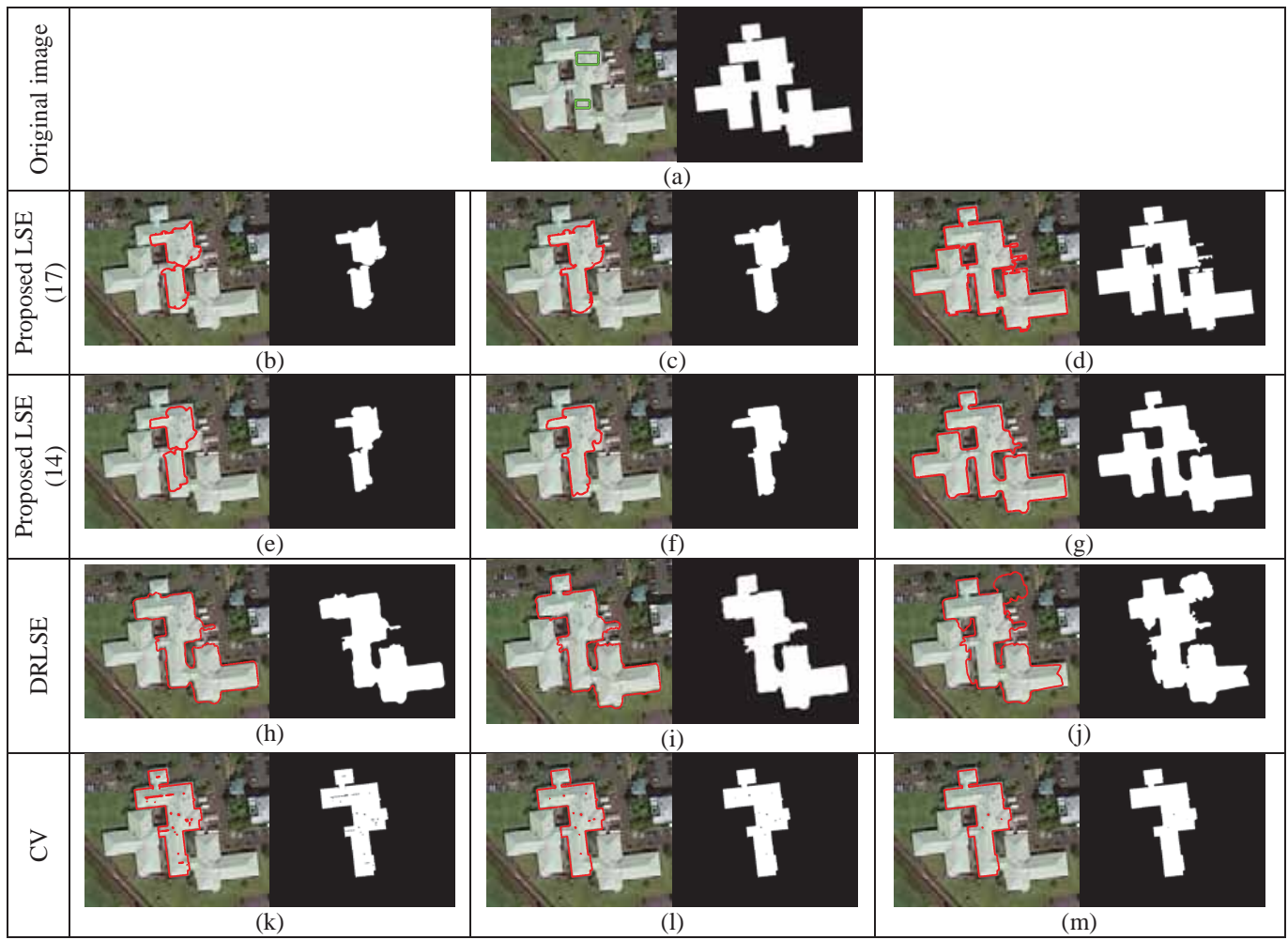

Fig. 7. Illustration of the merging of ZLCs of the LSEs (17), (14), DRLSE, and CV for single building roof extraction. (a) Left panel: the original image (i.e., building_2) with two initial ZLCs. Right panel: the ground truth. (b)-(d) Left panel in each pair: ZLCs of LSE (17) at iterations 15, 18, and 82, respectively. Right panel in each pair: corresponding binary results. (e)-(g) The ZLCs and binary results of LSE (14) at iterations 18, 24, and 148, respectively. (h)-(j) Results of DRLSE with different parameters used. $\alpha = 2$ and $\sigma = 5$ in (h), $\alpha = 2.5$ and $\sigma = 4$ in (i), and $\alpha = 3$ and $\sigma = 4$ in (j). (k)-(m) Results of CV with $\mu = 0.1$, 0.3, and 0.5, respectively.

### 2) Experiment 2: Airport Runway Extraction

This section aims to confirm the capabilities of the proposed LSEs (14) and (17) in 1) extracting desired objects with various geometric shapes from complicated backgrounds and 2) handling occlusions or discontinuities caused by the intensity variations.

Fig. 8 shows the results of all the LSEs for runway extraction from airport_1. The original image with initial ZLCs and the ground truth are shown in the left and right panels of Fig. 8(a), respectively. The runways in airport_1 are elongated features due to the relatively lower spatial resolution (see Table II), which often causes blurred edges and thus poses a great challenge for edge-based LSEs. It should be noted that ZLCs generally cannot pass through occlusions or discontinuities automatically. To obtain more complete and accurate airport runways, we set seven initial ZLCs manually at different positions in this experiment. The final ZLCs and corresponding binary results of LSEs (17), (14), DRLSE, and CV are displayed in Fig. 8(b)-(e), respectively. The parameters used for each LSE are given in Table III.

In general, in this experiment the performance of the region-based LSEs (i.e., the proposed (17) and CV) evidently surpass that of the edge-based ones (i.e., the proposed (14) and DRLSE). Despite some over-detection, region-based LSEs are able to extract the entire runways, as shown in Fig. 8(b) and (e), respectively. Edge-based ones, in contrast, just extract small part of the runways, as presented in Fig. 8(c) and (d), respectively. In addition, the ZLC of DRLSE passes through object boundaries and finally stops at the wrong boundaries. In this respect, the proposed edge-based LSE (14) has relatively better performance since no obvious boundary leakage happens, see Fig. 8(c).

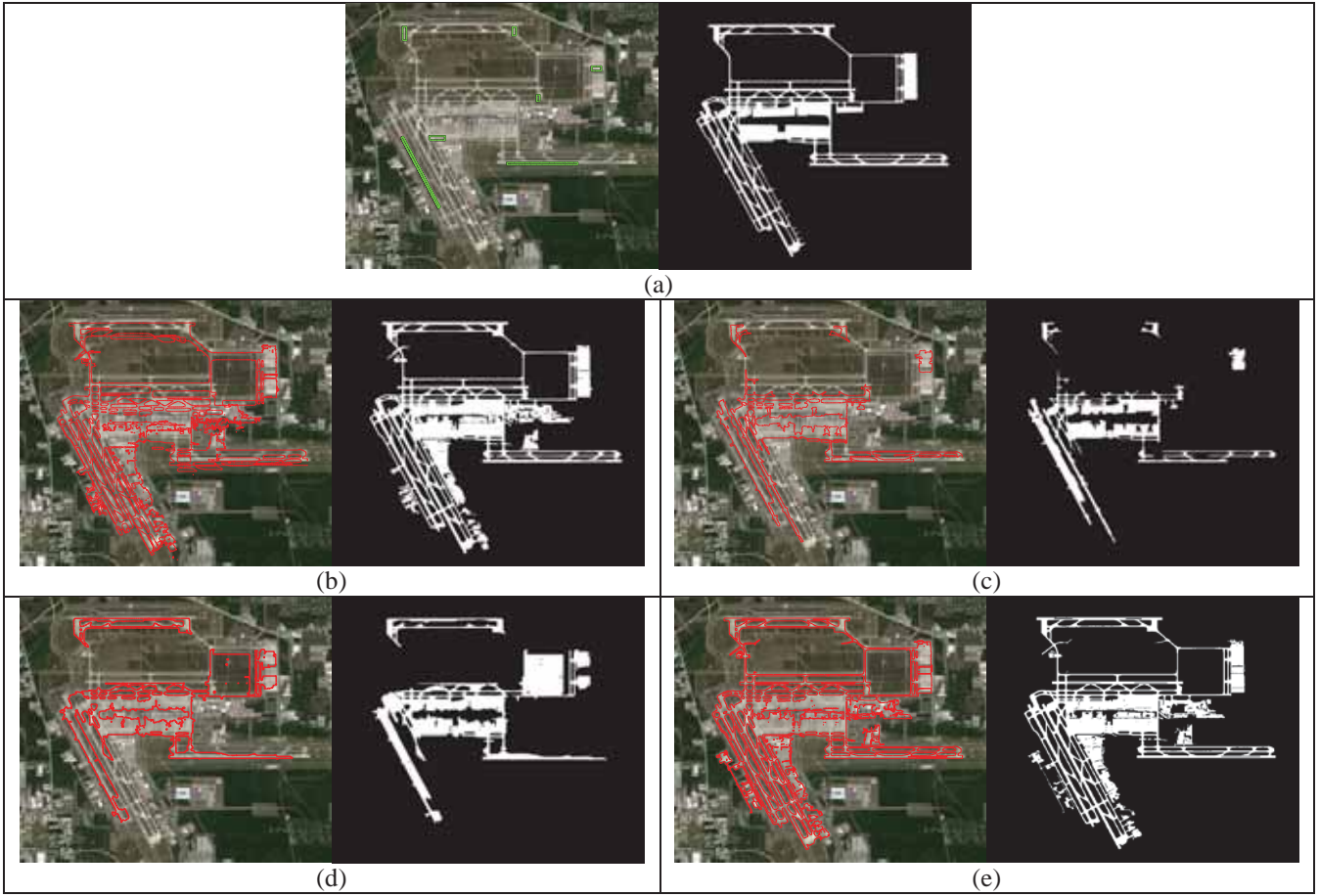

Fig. 8. Results of all the LSEs for runway extraction from airport_1. (a) Left panel: the original image. Right panel: the ground truth. (b)-(e) Left panel in each pair: final ZLCs of LSEs (17), (14), DRLSE, and CV, respectively. Right panel in each pair: corresponding binary results.

The results of all the LSEs for runway extraction from airport_2 are presented in Fig. 9. The spatial resolution of airport_2 is higher than that of airport_1 (see Table II). Thus, most runways in airport_2 appear wide and homogeneous. However, there are still some discontinuities on the runways due to the intensity variations. In this experiment we initialize six ZLCs at different positions to ensure that the whole runway can be extracted. The original image and ground truth are shown in the left and right panels of Fig. 9(a), respectively. The final results of LSEs (17), (14), DRLSE, and CV are illustrated in Fig. 9(b)-(e), respectively. Overall, despite some over-detection, region-based LSEs (i.e., (17) and CV) perform better than edge-based LSEs (i.e., (14) and DRLSE). As can be seen, region-based LSEs are able to extract almost all the airport runways, whereas the edge-based ones fail to extract the whole object. For DRLSE, it extracts more objects generally at the expense of boundary leakage, as shown in Fig. 9(d).

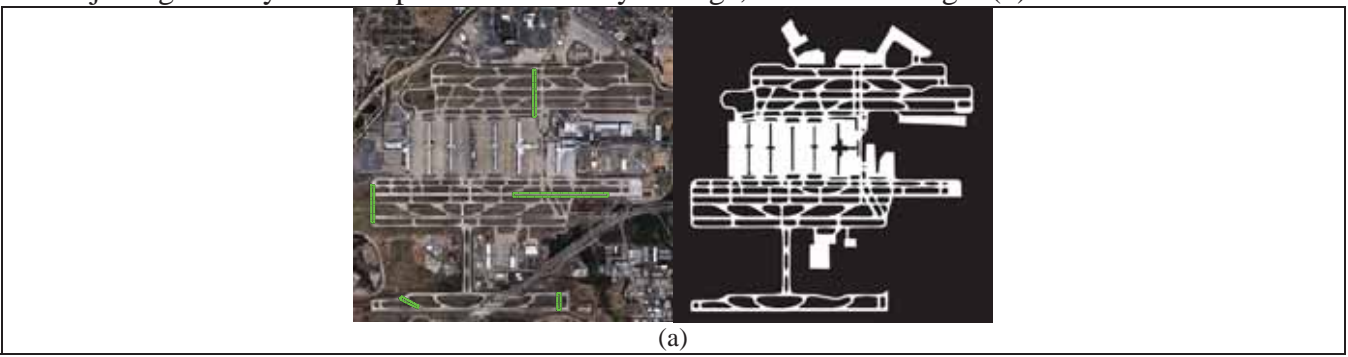

(a)

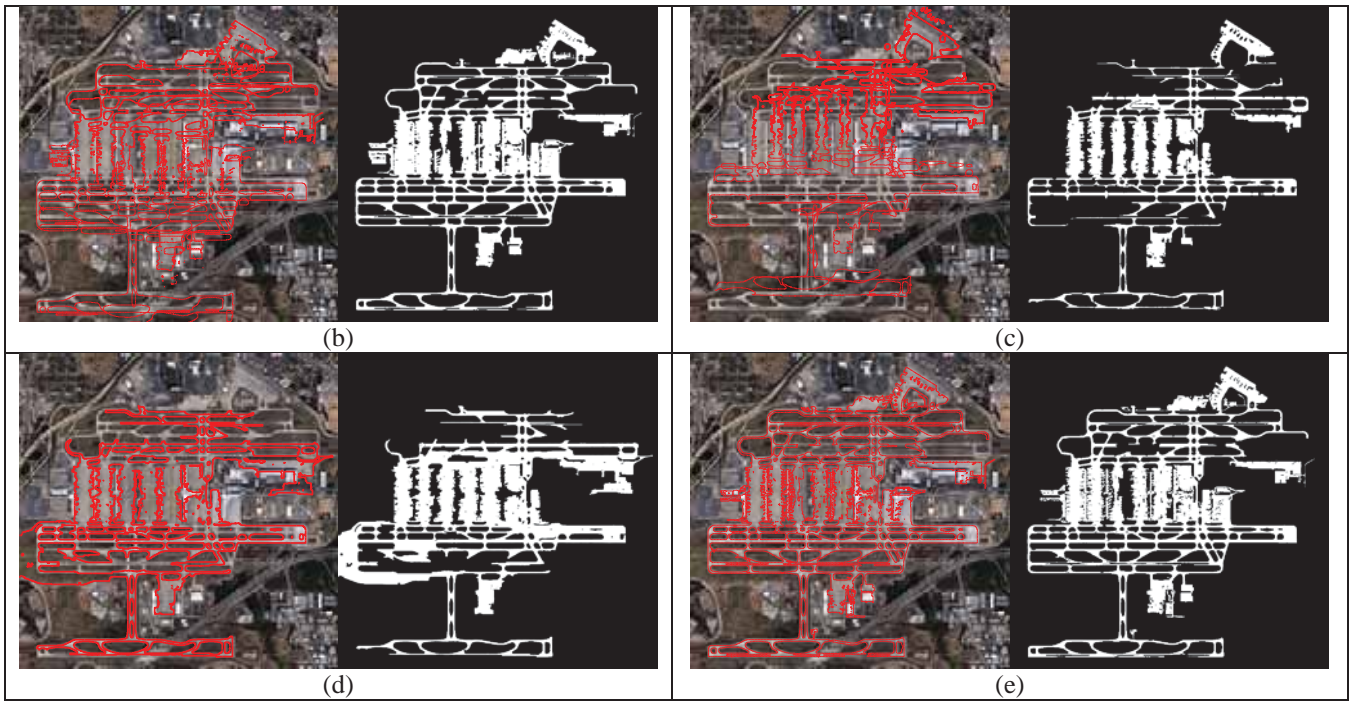

Fig. 9. Results of LSEs for runway extraction from airport_2. (a) Left panel: the original image. Right panel: the ground truth. (b)-(e) Results of the LSEs (17), (14), DRLSE, and CV, respectively.

### 3) Experiment 3: Road Network Extraction

The goal of this experiment is mainly to demonstrate the capability of the proposed LSEs (14) and (17) to deal with the roads with 1) discontinuities or occlusions caused by surrounding objects (e.g., trees and buildings) and 2) intensity variations caused by the use of different construction materials.

The results of all the LSEs for road network extraction are shown in Fig. 10. Similar to previous experiments, we initialize six ZLCs at different positions on the road to overcome the challenges caused by discontinuities or occlusions and ensure that the entire road can be extracted. The original image with initial ZLCs and the corresponding reference data are presented in Fig. 10(a). Note that two main challenging road segments are indicated by the yellow and red arrows, respectively. The former area indicated by the yellow arrow is surrounded by a number of trees and its continuity is broken. In comparison, the latter area indicated by the red arrow is occluded just by a couple of trees nearby. The road extraction results of LSEs (17), (14), DRLSE, and CV are shown in Fig. 10(b)-(e), respectively.

As can be seen in Fig. 10(c), the proposed edge-based LSE (14) has the best performance among all LSEs. It is capable of extracting all the road networks except the broken areas. As the same type of LSE, the performance of DRLSE, in contrast, is not as good as the proposed LSE (14). Examining Fig. 10(d), it just extracts small part of the road networks and boundary leakage also occurs at some places. Similar to LSE (14), region-based ones, i.e. the proposed (17) and CV, also have favorable performance. They not only extract the desired road networks, but they also extract those road segments that actually belong to private houses, which may be useful in some other specific applications. Note that they also have relatively poor performance at the place indicated by the yellow arrow.

In addition, we note that it is very difficult to determine the optimal parameters for DRLSE in the experiment of road extraction. Fig. 11 shows the results of DRLSE with different parameters used (see Table IV). According to [51], we fix $\lambda = 5.0$, $\mu = 0.04$, $\Delta t = 5.0$, and $c_0 = 2$ for DRLSE throughout this experiment. In the beginning, we set $\alpha = 3$, $\sigma = 1.5$, and the template size (TS) 9×9 for the Gaussian kernel. It converges at 2589th iteration and just extracts small part of the road, as displayed in Fig. 11(a). Then, we use a larger scale parameter $\sigma = 2$ to make the original image smoother and it indeed extracts more objects, as shown in Fig. 11(b). However, the boundary leakage arises at approximately iteration 4300. Based on this parameter tuning, we change the TS to 7×7. Finally, it is able to extract only small parts of the road; however, no boundary leakage happens, as presented in Fig. 11(c). According to the repeated parameter tuning, we empirically found that a larger value of $\alpha$ can expedite the evolution of ZLCs; however, it generally causes boundary leakage. As shown in Fig. 11(d) and (e), when we use a larger value of $\alpha = 4$, the ZLCs pass

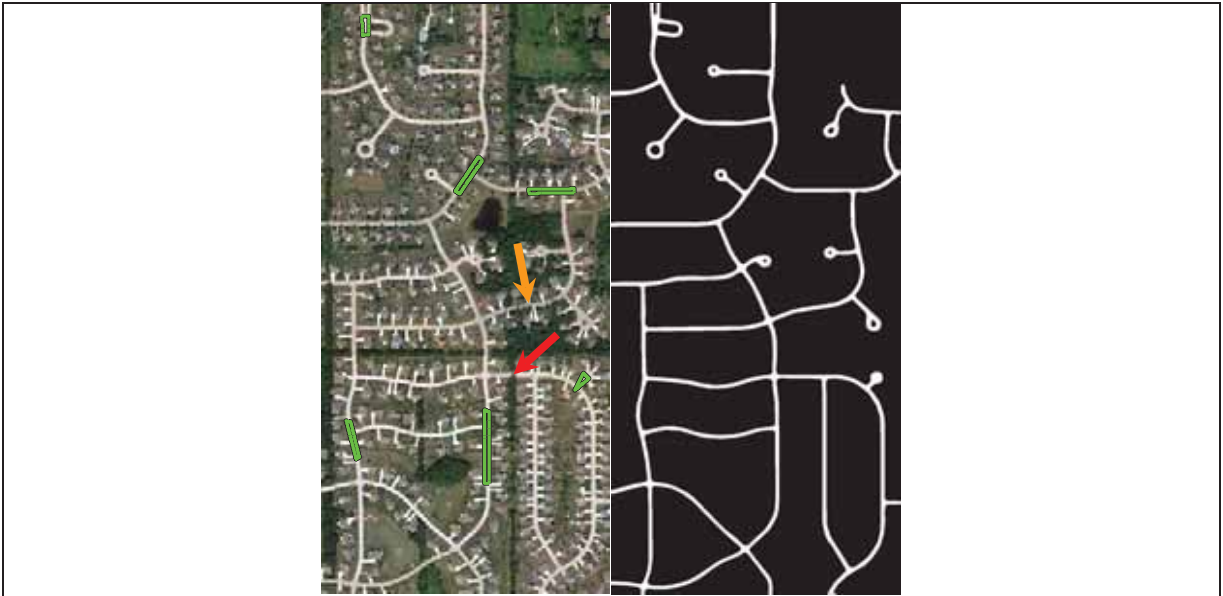

(a)

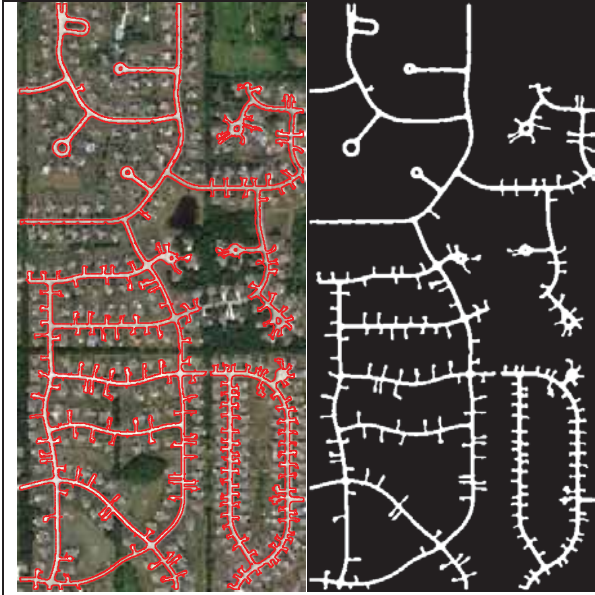

(b)

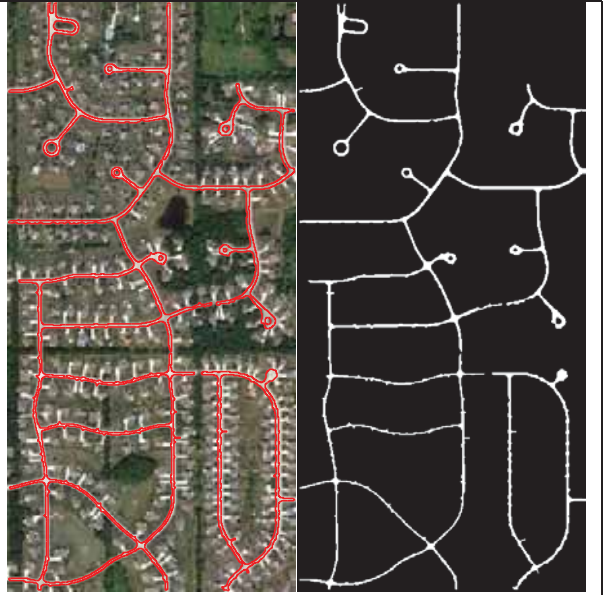

(c)

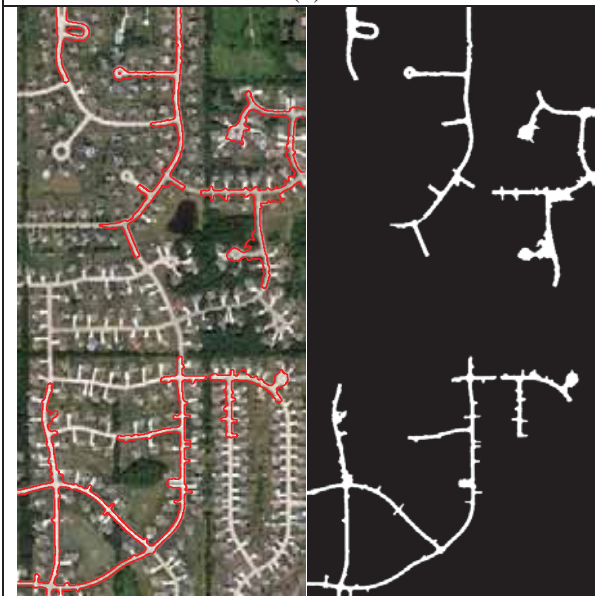

(d)

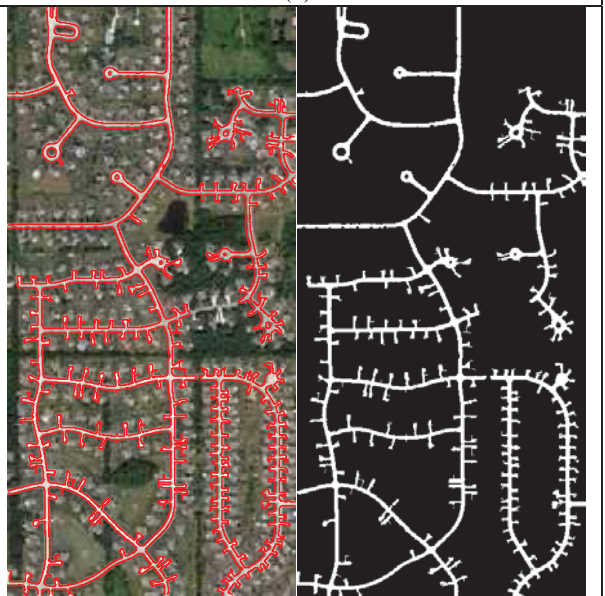

(e)

Fig. 10. Results of LSEs for road network extraction. (a) Left panel: the original image. Right panel: the ground truth. (b)-(e) Results of LSEs (17), (14), DRLSE, and CV, respectively. Left panel in each pair presents the final ZLCs of each LSE, whereas right panel shows the corresponding binary results.

TABLE IV
PARAMETERS USED BY DRLSE FOR ROAD EXTRACTION AND IMPORTANT ONES ARE HIGHLIGHTED
(UNIT FOR $\Delta t$: SECOND )

| Fig. 11 | Parameters |
|---|---|
| (a) | $\alpha = 3.0, \sigma = 1.5$ and TS 9×9 with fixed $\Delta t = 5, \lambda = 5, \mu = 0.04,$ and $c_0 = 2$ |
| (b) | $\alpha = 3.0, \sigma = 2.0$ and TS 9×9 with fixed $\Delta t = 5, \lambda = 5, \mu = 0.04,$ and $c_0 = 2$ |
| (c) | $\alpha = 3.0, \sigma = 2.0$ and TS 7×7 with fixed $\Delta t = 5, \lambda = 5, \mu = 0.04,$ and $c_0 = 2$ |
| (d) | $\alpha = 4.0, \sigma = 1.5$ and TS 9×9 with fixed $\Delta t = 5, \lambda = 5, \mu = 0.04,$ and $c_0 = 2$ |
| (e) | $\alpha = 4.0, \sigma = 1.5$ and TS 7×7 with fixed $\Delta t = 5, \lambda = 5, \mu = 0.04,$ and $c_0 = 2$ |
| (f) | $\alpha = 3.5, \sigma = 1.5$ and TS 9×9 with fixed $\Delta t = 5, \lambda = 5, \mu = 0.04,$ and $c_0 = 2$ |
| (g) | $\alpha = 3.5, \sigma = 1.5$ and TS 7×7 with fixed $\Delta t = 5, \lambda = 5, \mu = 0.04,$ and $c_0 = 2$ |
| (h) | $\alpha = 3.8, \sigma = 1.6$ and TS 7×7 with fixed $\Delta t = 5, \lambda = 5, \mu = 0.04,$ and $c_0 = 2$ |
| (i) | $\alpha = 3.8, \sigma = 1.5$ and TS 9×9 with fixed $\Delta t = 5, \lambda = 5, \mu = 0.04,$ and $c_0 = 2$ |
| (j) | $\alpha = 3.8, \sigma = 1.6$ and TS 9×9 with fixed $\Delta t = 5, \lambda = 5, \mu = 0.04,$ and $c_0 = 2$ |

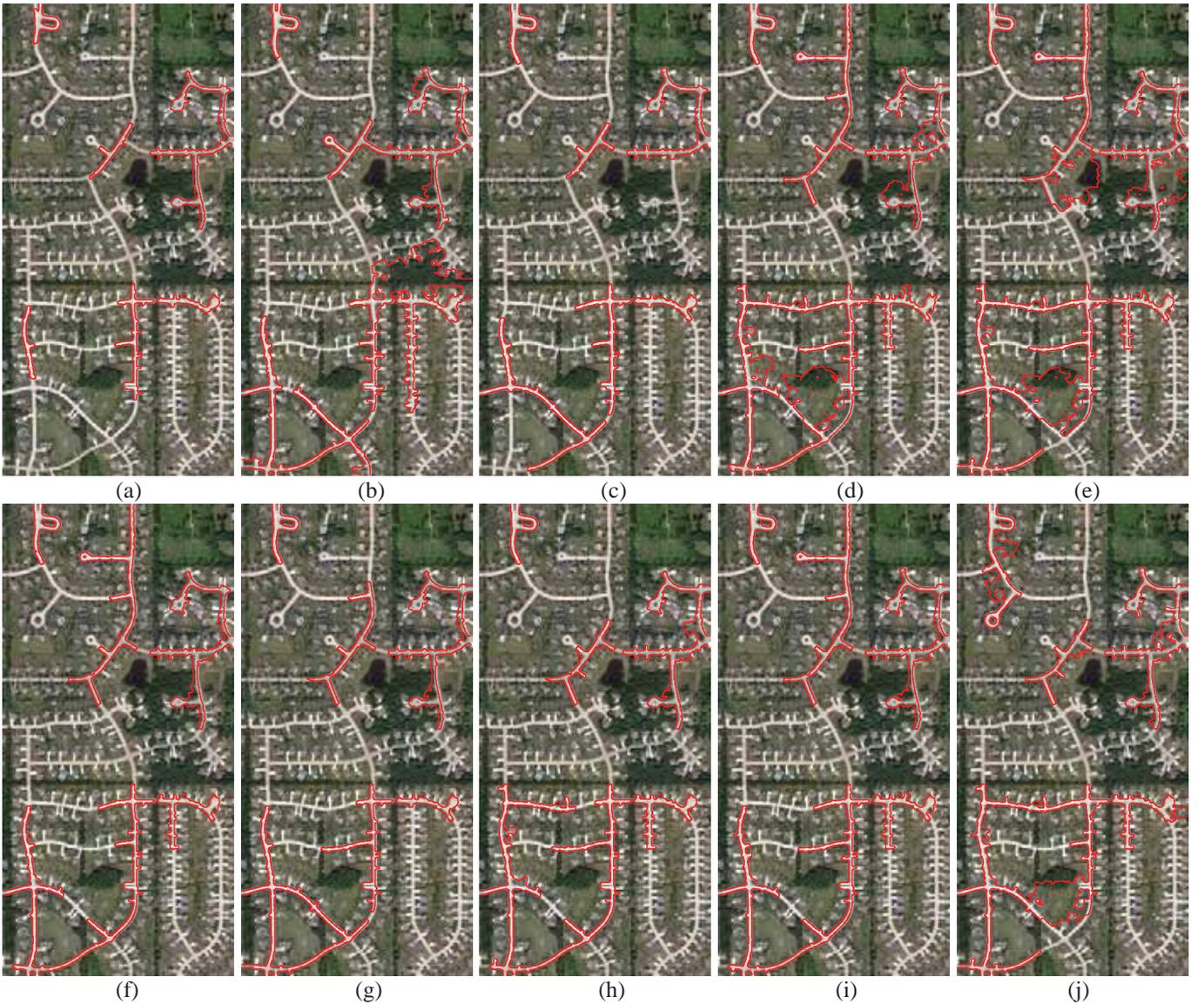

Fig. 11. Results of DRLSE for road network extraction with different parameters. See Table IV for parameters.

through the road boundaries at some places and extract nearby undesired objects. In addition, larger values of scale parameter $\sigma$ and TS often results in over-smoothing of the original image, which indirectly leads to boundary leakage. Thus, for the application of DRLSE, it is necessary to make a compromise between the efficiency and accuracy.

To avoid boundary leakage, we instead use a set of smaller values of $\alpha$ in Fig. 11(f) to (j). However, with none of the parameters it is feasible for DRLSE to extract the whole road networks. Moreover, it is worth mentioning that the result of DRLSE with $\alpha = 3.5$ appears to be identical to that of $\alpha = 3.8$, as shown in Fig. 11(f) and (i). After trial and error, we finally seek out a relatively satisfactory result among all the tests, namely, the one demonstrated in Fig. 10(d).

From the qualitative point of view, the performance of the proposed edge-based LSE (14) is much more favorable than that of DRLSE in all the experiments. In comparison, no obvious difference can be found between the proposed region-based LSE (17) and CV except the experiment of building_2. Compared with LSE (14), LSE (17) is able to obtain more accurate and complete object boundaries and it is more robust to intensity variations, see Figs. 8 and 9. However, in some cases LSE (14) performed better than LSE (17). For instance, in the experiment of road extraction in Fig. 10, LSE (14) failed to extract the roads that belong to the private houses since it is sensitive to intensity variations, which however makes its performance the best among all the LSEs. In addition, it should be noted that due to the convolution of original images with the Gaussian kernel, object boundaries extracted by edge-based LSEs (i.e., (14) and DRLSE) are smoother than those obtained by region-based LSEs, i.e., (17) and CV in all the experiments.

*C. Quantitative Evaluation*

In previous section, the proposed LSEs (14) and (17) were qualitatively compared with the existing popular LSEs, i.e., DRLSE and CV for man-made object extraction. In this section, we evaluate their performance quantitatively using four widely used indices, i.e., completeness, correctness, quality, and ratio of CPU time, as presented in Fig. 12(a)-(d).

Regarding the completeness, the proposed region-based LSE (17) performs better than other LSEs. Specifically, LSE (17) clearly outperforms CV in extracting building_2, as shown in Fig. 12(a). In addition to that, it is comparable to CV for the extraction of other man-made objects. Compared with DRLSE, the proposed edge-based LSE (14) performs better in extracting building_1, building_2, and road. For the extraction of airport_1, airport_2, and road, region-based LSEs (17) and CV exceed edge-based ones (14) and DRLSE significantly.

With respect to the correctness, the proposed edge-based LSE (14) exhibits obvious advantage over its competitors except for the extraction of building_1, as shown in Fig. 12(b). In addition, LSE (17) is comparable to CV in all experiments.

The quality of all the LSEs for man-made object extraction is plotted in Fig. 12(c). As seen clearly, the proposed LSE (17) surpasses other rivals in extracting the first four types of man-made objects. Also, the proposed LSE (14) outperforms its competitor (i.e., DRLSE) in extracting the last four types of objects. Additionally, it has the best performance in the experiment of road extraction compared with other LSEs. In contrast, the quality of CV is not very high for extraction of building_2 and DRLSE has not high quality for airport_1 and road.

Finally, the ratio of CPU time is presented in Fig. 12(d). The CPU time consumed by each LSE was recorded in Table V. Note that for fair comparison, we just recorded the CPU time of the best performance for each LSE. As can be seen clearly, the proposed LSEs are much more efficient than state-of-the-art ones. More precisely, the proposed region-based LSE (17) is at least 20 times faster than CV and the proposed edge-based LSE (14) is at least 35 times faster than DRLSE. For the comparison between the state-of-the-art LSEs, CV is almost 3 times faster than DRLSE, whereas no obvious difference between the proposed LSEs (14) and (17) can be found. By contrast, DRLSE is the slowest among all the LSEs.

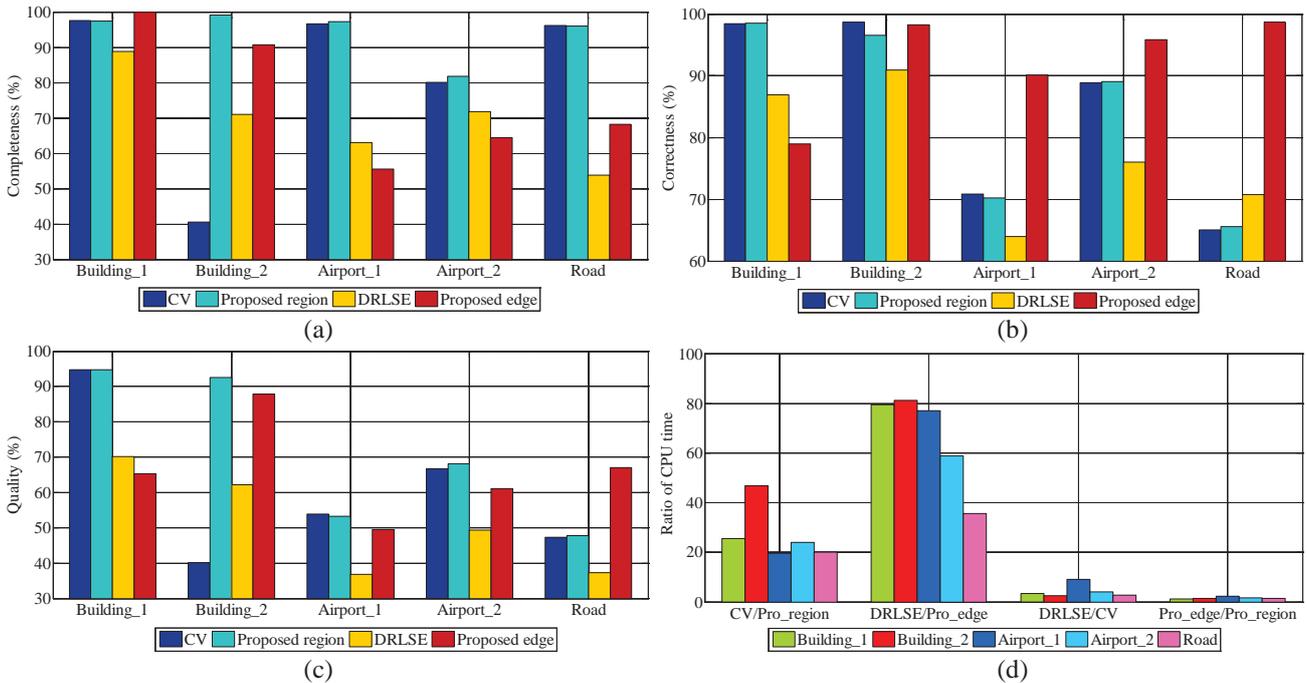

Fig. 12. Quantitative evaluation of LSEs (14), (17), DRLSE, and CV for man-made object extraction. (a) Completeness. (b) Correctness. (c) Quality. (d) Ratio of CPU time.

TABLE V
THE BEST CPU TIME FOR EACH LSE (UNIT: SECOND)

| Experiments | CV | Proposed region-based LSE (17) | DRLSE | Proposed edge-based LSE (14) |
|---|---|---|---|---|
| Building_1 | 50.16 | 1.96 | 175.48 | 2.21 |
| Building_2 | 75.46 | 1.61 | 188.59 | 2.32 |
| Airport_1 | 938.06 | 47.8 | 8635.78 | 111.92 |
| Airport_2 | 4792.67 | 198.78 | 19467.42 | 330.05 |
| Road | 362.52 | 18.07 | 959.84 | 26.90 |

### D. Parameter Analysis

Experiments show that the proposed LSEs (14) and (17) are more robust to parameter variation and the optimal parameters are much more readily available than other two state-of-the-art LSEs, i.e., DRLSE and CV.

As shown in Table III, there are various kinds of parameters in the LSEs, which need to be tuned before they can be used in practice. We have highlighted the most important ones for each LSE. Generally, we fix time step as $\Delta t = 15$ for a rapid convergence rate of LSEs (14) and (17). Note that if time step is larger than 25, unstable results may arise in both the LSEs. In practical applications, we therefore just need to tune the parameter $\sigma$ for (17) and $\sigma_1$ for (14), respectively. The larger value of $\sigma$ in (17) is helpful for filtering out those small extraneous objects, whereas the larger value of $\sigma_1$ in (14) is used to remove the noises and make the original image smoother. As for CV, the optimal parameters can be determined after a small number of tests. Empirically, the value of the parameter $\mu$ should be less than 1.0, whereas the time step should not be larger than 2.0. In contrast, DRLSE is very sensitive to the parameter variation. In particular, the small changes of the parameter values of $\alpha$, $\sigma$ or TS may cause serious boundary leakages, as presented in Fig. 11. In all experiments, we fixed other parameters for DRLSE according to the recommended values in [51]. Despite that, it is still challenging to seek out the optimal parameters for DRLSE in specific engineering applications. In current stage, a reliable and feasible way to solve this problem is to use trial and error.

## V. DISCUSSION

### A. Multiclass Object Extraction

For object extraction, the algorithms that only make use of the geometric shape, or spectral information, or some other special features may restrict their applications because different objects have their own intrinsic features. From a generic point of view, LSEs are more advantageous. Experiments show that they are capable of extracting most man-made objects. That is because they take advantage of the information of the intensity differences between the desired objects and the nearby undesired objects, e.g., the *edge function* $g(I)$ and $c^+$ and $c^-$ on each side of the ZLC, rather than geometric shapes, or multispectral information, or other characteristics. In addition, the involvement of the appropriate human interaction makes the proposed methods robust and flexible. In this way, they are more generic and operational.

### B. Spectral Features of Man-Made Objects

Each object has its own distinctive spectral (e.g., color and hue) or radiometric property (e.g., brightness, intensity, and tone) in optical multispectral images. Thus, they can be recognized by using their spectral signatures. For instance, NDVI derived from R (0.655-0.690 um) and NIR bands (0.780-0.920 um) is often used to extract the vegetation information. However, the spectral difference at the range of visible (R, G, and B) and NIR bands may not be an effective characteristic to represent man-made objects. This can be validated by the spectral reflectance data, as shown in Fig. 13 below. This data is publicly available and can be downloaded from ASTER Spectral Library, at Jet Propulsion Laboratory (JPL) and California Institute of Technology (Caltech), NASA. Here, we just discuss the spectral reflectance data of the construction asphalt and concrete since they are the commonly used materials for man-made objects. For comparison, we also give the spectral reflectance of the deciduous tree. As shown clearly, the deciduous tree has larger spectral difference between the R at NIR ranges. That is the reason why NDVI is effective for describing vegetation information. In contrast, no obvious spectral difference can be found for both construction asphalt and concrete throughout all the bands (i.e., B, G, R, and NIR band).

Generally, it is accepted that the asphalt and concrete appear to be darker and brighter parts in high spatial resolution multispectral images, respectively. The intensity contrast between the desired objects and the unwanted objects nearby, i.e., the radiometric differences are dominant for man-made object extraction. In addition, it should be noted that only four spectral bands (i.e., R, G, B, and NIR) are currently available for high spatial resolution images offered by sensors such as Ikonos, Quickbird, Pleiades-1, and Geoeye-1.

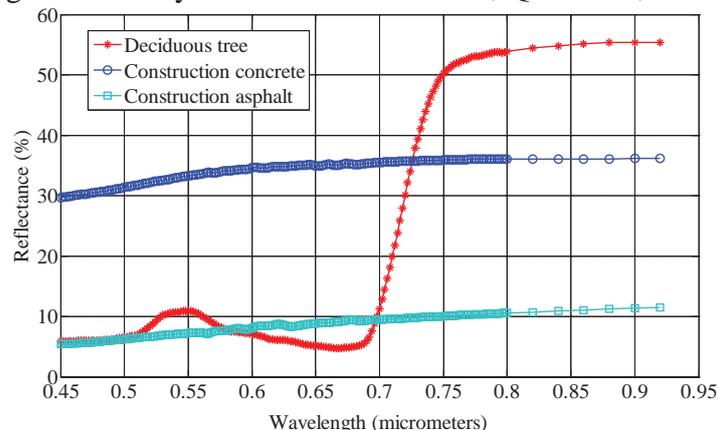

Fig. 13. The spectral reflectance of the deciduous tree, construction asphalt, and construction concrete at the range of visible (R, G, and B) and NIR bands.

Based on the above analysis, we conclude that man-made objects are mainly characterized by the radiometric difference (i.e., brightness, intensity, and tone) from neighboring objects instead of their spectral differences (i.e., color and hue) at the range of visible and NIR bands. This is the reason why we choose the intensity information instead of the spectral information for man-made object extraction.

### C. Limitations of the Proposed Level Set Evolutions

The main limitations of the proposed LSEs (14) and (17) are twofold. First, they require human interaction to set the initial ZLC, which means that they are a semiautomatic method. Currently, it is still a challenging task to make them fully automated. This is because the initial ZLC cannot be determined automatically. However, the degree of automation of the proposed LSEs is acceptable since only limited human input is required. In 2004, Baltsavias [10] mentioned that fully automated object extraction is an open problem. In 2008, Mayer [23] pointed out that it is very important and helpful to consider efficient human interaction in devising a practical object extraction system. In 2010, Blaschke [22] also mentioned that automated object recognition is an end goal. Second, it is currently not easy for us to estimate the scale parameter values such as $\sigma$ for the region-based LSE (17) and $\sigma_1$ and $\sigma_2$ for the edge-based LSE (14) adaptively. That is because it is not easy to obtain the accurate noise levels of the original images in advance.

### D. Future Work

As discussed above, there is large room to increase the degree of automation of the proposed LSEs. To this end, we intend to use some inherent geometric features (e.g., corners for building roofs and straight lines for roads or runways) of the desired objects to automate the initialization of ZLC. In the meantime, to obtain the scale parameter values for the proposed LSEs automatically, we intend to estimate the noise levels of the original images in advance. In this study we primarily focused on extracting man-made objects without considering natural objects. Thus, future research can also be directed at extraction of natural objects such as tree crown, surface water body, and shoreline. Moreover, for some specific applications such as agriculture mapping, it would be promising to integrate texture features like in [66] into LSEs. In addition, it would be promising to merge the two proposed LSEs together to handle multiple types of objects from one scene simultaneously.

## VI. Conclusion

In this study, we proposed two fast level set evolution (LSE) formulations: the edge-based LSE (14) and the region-based one (17), in which the traditional regularization term is replaced by a Gaussian kernel in a separate step and it is mathematically sound to do that. This makes it possible to use a relatively larger time step in their numerical schemes, thus expediting their convergence considerably. Moreover, a new data term was devised for the proposed region-based LSE (17) that makes it more robust to the initial position of ZLC, the sign of LSF, and the image noise compared with the proposed edge-based LSE (14).

Experiments show that the LSEs are capable of extracting most man-made objects from remote sensing images by just using the gradient or region statistics (i.e., intensity mean) instead of the typically used geometric shapes and multispectral signatures. However, compared with state-of-the-art LSEs in extracting building roofs, road networks, and airport runways, the proposed LSEs are computationally much more efficient while achieving better performance. Most importantly, they have much fewer input parameters, which make them advantageous in practical applications.

## Acknowledgment

The authors would like to thank Dr. Z.Z. Liu for the valuable conversations. Also, they wish to thank all the unknown reviewers for their detailed and insightful comments and suggestions, which improved very much the quality of the paper.

## REFERENCES

[1] C. S. Wu and A. T. Murray, "Estimating impervious surface distribution by spectral mixture analysis," *Remote. Sens. Environ.,* vol. 84, pp. 493-505, Apr. 2003.

[2] S. L. Powell, W. B. Cohen, Z. Yang, J. D. Pierce, and M. Alberti, "Quantification of impervious surface in the Snohomish Water Resources Inventory Area of Western Washington from 1972-2006," *Remote. Sens. Environ.,* vol. 112, pp. 1895-1908, Apr. 2008.

[3] Q. H. Weng, "Remote sensing of impervious surfaces in the urban areas: Requirements, methods, and trends," *Remote. Sens. Environ.,* vol. 117, pp. 34-49, Feb. 2012.

[4] R. Duca and F. Del Frate, "Hyperspectral and Multiangle CHRIS-PROBA Images for the Generation of Land Cover Maps," *IEEE Trans. Geosci. Remote Sens.,* vol. 46, pp. 2857-2866, Oct. 2008.

[5] R. J. Dekker, "Texture analysis and classification of ERS SAR images for map updating of urban areas in the Netherlands," *IEEE Trans. Geosci. Remote Sens.,* vol. 41, pp. 1950-1958, Sep. 2003.

[6] J. B. Mena, "State of the art on automatic road extraction for GIS update: a novel classification," *Pattern Recognit. Lett.,* vol. 24, pp. 3037-3058, Dec. 2003.

[7] S. E. Park, Y. Yamaguchi, and D. J. Kim, "Polarimetric SAR remote sensing of the 2011 Tohoku earthquake using ALOS/PALSAR," *Remote. Sens. Environ.,* vol. 132, pp. 212-220, May. 2013.

[8] X. H. Tong, Z. H. Hong, S. J. Liu, X. Zhang, H. Xie, Z. Y. Li, S. L. Yang, W. A. Wang, and F. Bao, "Building-damage detection using pre- and post-seismic high-resolution satellite stereo imagery: A case study of the May 2008 Wenchuan earthquake," *ISPRS J. Photogramm. Remote Sens.,* vol. 68, pp. 13-27, Mar. 2012.

[9] O. Aytekin, U. Zongur, and U. Halici, "Texture-Based Airport Runway Detection," *IEEE Geosci. Remote Sens. Lett.,* vol. 10, pp. 471-475, May. 2013.

[10] E. P. Baltsavias, "Object extraction and revision by image analysis using existing geodata and knowledge: current status and steps towards operational systems," *ISPRS J. Photogramm. Remote Sens.,* vol. 58, pp. 129-151, Jan. 2004.

[11] K. Hedman, U. Stilla, G. Lisini, and P. Gamba, "Road Network Extraction in VHR SAR Images of Urban and Suburban Areas by Means of Class-Aided Feature-Level Fusion," *Ieee T Geosci Remote,* vol. 48, pp. 1294-1296, Mar. 2010.

[12] J. Hu, A. Razdan, J. C. Femiani, M. Cui, and P. Wonka, "Road network extraction and intersection detection from aerial images by tracking road footprints," *Ieee T Geosci Remote,* vol. 45, pp. 4144-4157, Dec. 2007.

[13] W. Shi, Z. Miao, and J. Debayle, "An Integrated Method for Urban Main-Road Centerline Extraction From Optical Remotely Sensed Imagery," *IEEE Trans. Geosci. Remote Sens.,* vol. PP, pp. 1-14, 2013.

[14] M. Ortner, X. Descombes, and J. Zerubia, "A marked point process of rectangles and segments for automatic analysis of Digital Elevation Models," *IEEE Trans. Pattern Anal. Mach. Intell.,* vol. 30, pp. 105-119, Jan. 2008.

[15] S. Noronha and R. Nevatia, "Detection and modeling of buildings from multiple aerial images," *IEEE Trans. Pattern Anal. Mach. Intell.,* vol. 23, pp. 501-518, May. 2001.

[16] B. Sirmacek and C. Unsalan, "A Probabilistic Framework to Detect Buildings in Aerial and Satellite Images," *IEEE Trans. Geosci. Remote Sens.,* vol. 49, pp. 211-221, Jan. 2011.

[17] M. Cote and P. Saeedi, "Automatic Rooftop Extraction in Nadir Aerial Imagery of Suburban Regions Using Corners and Variational Level Set Evolution," *IEEE Trans. Geosci. Remote Sens.,* vol. 51, pp. 313-328, Jan. 2013.

[18] A. O. Ok, C. Senaras, and B. Yuksel, "Automated Detection of Arbitrarily Shaped Buildings in Complex Environments From Monocular VHR Optical Satellite Imagery," *IEEE Trans. Geosci. Remote Sens.,* vol. 51, pp. 1701-1717, Mar. 2013.

[19] J. C. Harsanyi and C. I. Chang, "Hyperspectral Image Classification and Dimensionality Reduction - an Orthogonal Subspace Projection Approach," *IEEE Trans. Geosci. Remote Sens.,* vol. 32, pp. 779-785, Jul. 1994.

[20] D. Azar, R. Engstrom, J. Graesser, and J. Comenetz, "Generation of fine-scale population layers using multi-resolution satellite imagery and geospatial data," *Remote. Sens. Environ.,* vol. 130, pp. 219-232, Mar 15. 2013.

[21] D. Manolakis, C. Siracusa, and G. Shaw, "Hyperspectral subpixel target detection using the linear mixing model," *IEEE Trans. Geosci. Remote Sens.,* vol. 39, pp. 1392-1409, Jul. 2001.

[22] T. Blaschke, "Object based image analysis for remote sensing," *ISPRS J. Photogramm. Remote Sens.,* vol. 65, pp. 2-16, Jan. 2010.

[23] H. Mayer, "Object extraction in photogrammetric computer vision," *ISPRS J. Photogramm. Remote Sens.,* vol. 63, pp. 213-222, Mar. 2008.

[24] N. H. Younan, S. Aksoy, and R. L. King, "Foreword to the Special Issue on Pattern Recognition in Remote Sensing," *IEEE J. Sel. Topics Appl. Earth Observ.,* vol. 5, pp. 1331-1334, Oct. 2012.

[25] L. Fang, M. Wang, D. Li, and J. Pan, "CPU/GPU near real-time preprocessing for ZY-3 satellite images: Relative radiometric correction, MTF compensation, and geocorrection," *ISPRS J. Photogramm. Remote Sens.,* vol. 87, pp. 229-240, 2014.

[26] C. Brenner, "Building reconstruction from images and laser scanning," *Int. J. Appl. Earth Obs. Geoinf.,* vol. 6, pp. 187-198, Mar. 2005.

[27] P. Perona and J. Malik, "Scale-Space and Edge-Detection Using Anisotropic Diffusion," *IEEE Trans. Pattern Anal. Mach. Intell.,* vol. 12, pp. 629-639, Jul. 1990.

[28] L. Alvarez, P. L. Lions, and J. M. Morel, "Image Selective Smoothing and Edge-Detection by Nonlinear Diffusion .2.," *Siam J. Numer. Anal.,* vol. 29, pp. 845-866, Jun. 1992.

[29] F. Tupin, B. Houshmand, and M. Datcu, "Road detection in dense urban areas using SAR imagery and the usefulness of multiple views," *Ieee T Geosci Remote,* vol. 40, pp. 2405-2414, Nov. 2002.

[30] B. K. Jeon, J. H. Jang, and K. S. Hong, "Road detection in spaceborne SAR images using a genetic algorithm," *Ieee T Geosci Remote,* vol. 40, pp. 22-29, Jan. 2002.

[31] W. Z. Shi and C. Q. Zhu, "The line segment match method for extracting road network from high-resolution satellite images," *Ieee T Geosci Remote,* vol. 40, pp. 511-514, Feb. 2002.

[32] M. Izadi and P. Saeedi, "Three-Dimensional Polygonal Building Model Estimation From Single Satellite Images," *IEEE Trans. Geosci. Remote Sens.,* vol. 50, pp. 2254-2272, Jun. 2012.

[33] F. Xu and Y. Q. Jin, "Automatic reconstruction of building objects from multiaspect meter-resolution SAR Images," *IEEE Trans. Geosci. Remote Sens.,* vol. 45, pp. 2336-2353, Jul. 2007.

[34] A. Katartzis and H. Sahli, "A stochastic framework for the identification of building rooftops using a single remote sensing image," *IEEE Trans. Geosci. Remote Sens.,* vol. 46, pp. 259-271, Jan. 2008.

[35] R. Stoica, X. Descombes, and J. Zerubia, "A Gibbs point process for road extraction from remotely sensed images," *Int. J. Comput. Vis.,* vol. 57, pp. 121-136, May. 2004.

[36] C. Lacoste, X. Descombes, and J. Zerubia, "Point processes for unsupervised line network extraction in remote sensing," *Ieee T Pattern Anal,* vol. 27, pp. 1568-1579, Oct. 2005.

[37] M. Ortner, X. Descombes, and J. Zerubia, "Building outline extraction from digital elevation models using marked point processes," *Int. J. Comput. Vis.,* vol. 72, pp. 107-132, Apr. 2007.

[38] O. Tournaire, M. Bredif, D. Boldo, and M. Durupt, "An efficient stochastic approach for building footprint extraction from digital elevation models," *ISPRS J. Photogramm. Remote Sens.,* vol. 65, pp. 317-327, Jul. 2010.

[39] M. Negri, P. Gamba, G. Lisini, and F. Tupin, "Junction-aware extraction and regularization of urban road networks in high-resolution, SAR images," *IEEE Trans. Geosci. Remote Sens.,* vol. 44, pp. 2962-2971, Oct. 2006.

[40] S. Das, T. T. Mirnalinee, and K. Varghese, "Use of Salient Features for the Design of a Multistage Framework to Extract Roads From High-Resolution Multispectral Satellite Images," *Ieee T Geosci Remote,* vol. 49, pp. 3906-3931, Oct. 2011.

[41] M. Awrangjeb, M. Ravanbakhsh, and C. S. Fraser, "Automatic detection of residential buildings using LIDAR data and multispectral imagery," *ISPRS J. Photogramm. Remote Sens.,* vol. 65, pp. 457-467, Sep. 2010.

[42] I. Sebari and D. C. He, "Automatic fuzzy object-based analysis of VHSR images for urban objects extraction," *ISPRS J. Photogramm. Remote Sens.,* vol. 79, pp. 171-184, May. 2013.

[43] X. H. Tong, X. F. Lin, T. T. Feng, H. Xie, S. J. Liu, Z. H. Hong, and P. Chen, "Use of shadows for detection of earthquake-induced collapsed buildings in high-resolution satellite imagery," *ISPRS J. Photogramm. Remote Sens.,* vol. 79, pp. 53-67, May. 2013.

[44] S. Osher and J. A. Sethian, "Fronts Propagating with Curvature-Dependent Speed - Algorithms Based on Hamilton-Jacobi Formulations," *J. Comput. Phys.,* vol. 79, pp. 12-49, Nov. 1988.

[45] R. P. Fedkiw, G. Sapiro, and C. W. Shu, "Shock capturing, level sets, and PDE based methods in computer vision and image processing: a review of Osher's contributions," *J. Comput. Phys.,* vol. 185, pp. 309-341, Mar 1. 2003.

[46] D. Cremers, M. Rousson, and R. Deriche, "A review of statistical approaches to level set segmentation: Integrating color, texture, motion and shape," *Int. J. Comput. Vis.,* vol. 72, pp. 195-215, Apr. 2007.

[47] N. P. van Dijk, K. Maute, M. Langelaar, and F. van Keulen, "Level-set methods for structural topology optimization: a review," *Struct. Multidiscip. Optim.,* vol. 48, pp. 437-472, Sep. 2013.

[48] M. B. Nielsen, O. Nilsson, A. Soderstrom, and K. Museth, "Out-of-core and compressed level set methods," *Acm T. Graphic.,* vol. 26, 2007.

[49] V. Caselles, F. Catte, T. Coll, and F. Dibos, "A Geometric Model for Active Contours in Image-Processing," *Numer. Math.,* vol. 66, pp. 1-31, Oct. 1993.

[50] V. Caselles, R. Kimmel, and G. Sapiro, "Geodesic active contours," *Int. J. Comput. Vis.,* vol. 22, pp. 61-79, Feb-Mar. 1997.

[51] C. M. Li, C. Y. Xu, C. F. Gui, and M. D. Fox, "Distance Regularized Level Set Evolution and Its Application to Image Segmentation," *IEEE Trans. Image Process.,* vol. 19, pp. 3243-3254, Dec. 2010.

[52] T. F. Chan and L. A. Vese, "Active contours without edges," *IEEE Trans. Image Process.,* vol. 10, pp. 266-277, Feb. 2001.

[53] S. C. Zhu and A. Yuille, "Region competition: Unifying snakes, region growing, and Bayes/MDL for multiband image segmentation," *IEEE Trans. Pattern Anal. Mach. Intell.,* vol. 18, pp. 884-900, Sep. 1996.

[54] K. Karantzalos and D. Argialas, "A Region-based Level Set Segmentation for Automatic Detection of Man-made Objects from Aerial and Satellite Images," *Photogramm. Eng. Remote Sens.,* vol. 75, pp. 667-677, Jun. 2009.

[55] A. Yezzi, A. Tsai, and A. Willsky, "A fully global approach to image segmentation via coupled curve evolution equations," *J. Vis. Commun. Image Represent.,* vol. 13, pp. 195-216, Mar-Jun. 2002.

[56] R. Kimmel, *Numerical geometry of images : theory, algorithms, and applications*. New York: Springer, 2004, pp. 51-54.

[57] S. Osher and R. Fedkiw, *Level Set Methods and Dynamic Implicit Surface*. New York, NY, USA: Springer-Verlag, 2002, pp. 123-124.

[58]  K. H. Zhang, L. Zhang, H. H. Song, and W. G. Zhou, "Active contours with selective local or global segmentation: A new formulation and level set method," *Image Vis. Comput.,* vol. 28, pp. 668-676, Apr. 2010.

[59]  Y. Shi and W. C. Karl, "A real-time algorithm for the approximation of level-set-based curve evolution," *IEEE Trans. Image Process.,* vol. 17, pp. 645-656, May. 2008.

[60]  B. Huang, H. Li, and X. Huang, "A level set method for oil slick segmentation in SAR images," *Int. J. Remote Sens.,* vol. 26, pp. 1145-1156, Mar. 2005.

[61]  W. Sun, M. Cetin, W. C. Thacker, T. M. Chin, and A. S. Willsky, "Variational approaches on discontinuity localization and field estimation in sea surface temperature and soil moisture," *IEEE Trans. Geosci. Remote Sens.,* vol. 44, pp. 336-350, Feb. 2006.

[62]  J. E. Ball and L. M. Bruce, "Level set hyperspectral image classification using best band analysis," *IEEE Trans. Geosci. Remote Sens.,* vol. 45, pp. 3022-3027, Oct. 2007.

[63]  K. Karantzalos and N. Paragios, "Recognition-Driven Two-Dimensional Competing Priors Toward Automatic and Accurate Building Detection," *IEEE Trans. Geosci. Remote Sens.,* vol. 47, pp. 133-144, Jan. 2009.

[64]  S. Ahmadi, M. J. V. Zoej, H. Ebadi, H. A. Moghaddam, and A. Mohammadzadeh, "Automatic urban building boundary extraction from high resolution aerial images using an innovative model of active contours," *Int. J. Appl. Earth Obs. Geoinf.,* vol. 12, pp. 150-157, Jun. 2010.

[65]  K. Kim and J. Shan, "Building roof modeling from airborne laser scanning data based on level set approach," *ISPRS J. Photogramm. Remote Sens.,* vol. 66, pp. 484-497, Jul. 2011.

[66]  Q. Wu and J. An, "An Active Contour Model Based on Texture Distribution for Extracting Inhomogeneous Insulators From Aerial Images," *IEEE Trans. Geosci. Remote Sens.,* vol. 49, pp. 3613-3626, 2014.

[67]  Y. M. Shu, J. Li, and G. Gomes, "Shoreline Extraction from RADARSAT-2 Intensity Imagery Using a Narrow Band Level Set Segmentation Approach," *Mar. Geod.,* vol. 33, pp. 187-203, 2010.

[68]  Y. Bazi, F. Melgani, and H. D. Al-Sharari, "Unsupervised Change Detection in Multispectral Remotely Sensed Imagery With Level Set Methods," *IEEE Trans. Geosci. Remote Sens.,* vol. 48, pp. 3178-3187, Aug. 2010.

[69]  T. Celik and K. K. Ma, "Multitemporal Image Change Detection Using Undecimated Discrete Wavelet Transform and Active Contours," *IEEE Trans. Geosci. Remote Sens.,* vol. 49, pp. 706-716, Feb. 2011.

[70]  M. I. Elbakary and K. M. Iftekharuddin, "Shadow Detection of Man-Made Buildings in High-Resolution Panchromatic Satellite Images," *IEEE Trans. Geosci. Remote Sens.,* vol. PP, pp. 1-13, 2013.

[71]  S. J. Tang, P. L. Dong, and B. P. Buckles, "Three-dimensional surface reconstruction of tree canopy from lidar point clouds using a region-based level set method," *Int. J. Remote Sens.,* vol. 34, pp. 1373-1385, 2013.

[72]  X. T. Niu, "A semi-automatic framework for highway extraction and vehicle detection based on a geometric deformable model," *ISPRS J. Photogramm. Remote Sens.,* vol. 61, pp. 170-186, Dec. 2006.

[73]  C. Shen, J. Fan, L. Pi, and F. Li, "Delineating lakes and enclosed islands in satellite imagery by geodesic active contour model," *Int. J. Remote Sens.,* vol. 27, pp. 5253-5268, Dec. 2006.

[74]  K. H. Zhang, L. Zhang, H. H. Song, and D. Zhang, "Reinitialization-Free Level Set Evolution via Reaction Diffusion," *IEEE Trans. Image Process.,* vol. 22, pp. 258-271, Jan. 2013.

[75]  F. Gibou and R. Fedkiw, "A fast hybrid k-means level set algorithm for segmentation," Stanford Univ., Stanford, CA, USA, Tech. Rep.2002.

[76]  L. Bertelli, S. Chandrasekaran, F. Gibou, and B. S. Manjunath, "On the Length and Area Regularization for Multiphase Level Set Segmentation," *Int. J. Comput. Vis.,* vol. 90, pp. 267-282, Dec. 2010.

[77]  S. Osher and R. Fedkiw, *Level Set Methods and Dynamic Implicit Surface*. New York, NY, USA: Springer-Verlag, 2002, pp. 18-22.

[78]  T. E. Chan, B. Y. Sandberg, and L. A. Vese, "Active contours without edges for vector-valued images," *J. Vis. Commun. Image Represent.,* vol. 11, pp. 130-141, Jun. 2000.

The MATLAB source code of the proposed edge-based LSE (14) is provided as follows:

```matlab
%%%%%%%%%%%%%%%%%%%%%%%%%%%%%%%%%%%%%%%%%%%%
%%% Zhongbin Li, Dept. LSGI, PolyU, Hong Kong%%%
%%% Email:lzbtongji@gmail.com              %%%
%%%%%%%%%%%%%%%%%%%%%%%%%%%%%%%%%%%%%%%%%%%%
clear all; close all; clc;
% Read images
[filename, pathname] = uigetfile ('*.jpg; *.jpeg; *.bmp; *.png; *.tif',
'Pick a file');
if ~filename
    return;
end
I0 = imread (strcat(pathname, filename));
[row, col, b] = size (I0);
if b > 1
    I = double(rgb2gray(I0));
elseif b == 1
    I = double(I0);
end

figure, imshow (I0,'border', 'tight'); hold on;
% Draw single zero level curve
% [junk px py] = roipoly;
% junk = junk + 0;
% Draw multiple zero level curves. This step can be further optimized.
[junk1 px1 py1] = roipoly;
junk1 = junk1 + 0;
[junk2 px2 py2] = roipoly;
junk2 = junk2 + 0;
[junk3 px3 py3] = roipoly;
junk3 = junk3 + 0;
[junk4 px4 py4] = roipoly;
junk4 = junk4 + 0;
[junk5 px5 py5] = roipoly;
junk5 = junk5 + 0;
junk = bitxor(bitxor(bitxor (bitxor (junk1, junk2), junk3), junk4), junk5);
junk(junk == 0) = -1;
phi = junk; contour (phi, [0 0], 'k', 'linewidth', 10);  contour (phi, [0
0], 'g', 'linewidth', 5);
figure, imshow (junk,'border', 'tight');

%I = addborder (I, 5, 255, 'inner'); % See it in file exchange MATLAB central
sigma_0 = 1;      % Scale parameter for Gaussian kernel
G_0 = fspecial('gaussian', 9, sigma_0);   % Size of the template should be
an odd number
Img_smooth = conv2(I, G_0, 'same'); % Smooth image by Gaussian convolution
[Ix, Iy] = gradient(Img_smooth);
f = Ix.^2+Iy.^2;
g = 1./(1 + f);   % Edge function.
figure, imshow (g);

sigma = 1; G = fspecial('gaussian', 9, sigma);
```

```matlab
delt = 15; % Time step
Iter =300; % Iteration number
tic;  % Timer
for n = 1: Iter
    [phi_x, phi_y] = gradient (phi);
    s = sqrt(phi_x.^2 + phi_y.^2);
    phi = phi + delt * g .* s;  % The formulation (14)

    if mod (n, 30)==0
        figure, imshow(I0,'border', 'tight');  hold on;
        contour (phi, [0 0], 'k', 'linewidth', 10);
        contour (phi, [0 0], 'r', 'linewidth', 5);
    end

    phi = conv2(phi, G, 'same');
    phi = (phi>=0) - (phi<0);

end
toc;
phi = (phi>=0) - (phi<0);
figure, imshow (phi,'border', 'tight');
```